\documentclass{IEEEtran}

\usepackage{mathtools}
\usepackage[utf8]{inputenc} 
\usepackage[T1]{fontenc}    
\usepackage{tcolorbox}
\usepackage[noend]{algpseudocode}
\usepackage[ruled,lined,linesnumbered]{algorithm2e}
\usepackage{bm}
\usepackage{stackrel}
\usepackage{subfigure}
\usepackage{epsf}
\usepackage{amsmath}
\usepackage{amssymb}
\usepackage{graphicx}
\usepackage{color}
\usepackage{cite}
\usepackage{multirow}
\usepackage{tabularx}
\usepackage{ifthen}
\usepackage{epstopdf}
\usepackage{wrapfig}

\input{epstopdf.sty}
\usepackage{times}

\input{epsf.sty}

\usepackage{subfigure}
\usepackage{caption}
\usepackage{booktabs} 
\usepackage{capt-of}
\usepackage{microtype}
\usepackage{hyperref}
\usepackage{xr}
\usepackage{tikz}

\newtheorem{theorem}{Theorem}

\newtheorem{corollary}[theorem]{Corollary}

\newtheorem{lemma}[theorem]{Lemma}

\newtheorem{definition}[theorem]{Definition}

\newtheorem{remark}[theorem]{Remark}

\newcommand{\bdmath}{\begin{dmath}}
	\newcommand{\edmath}{\end{dmath}}
\newcommand{\beq}{\begin{equation}}
	\newcommand{\eeq}{\end{equation}}
\newcommand{\bdm}{\begin{displaymath}}
	\newcommand{\edm}{\end{displaymath}}
\newcommand{\bea}{\begin{eqnarray}}
	\newcommand{\eea}{\end{eqnarray}}
\newcommand{\beal}{\beq \begin{array}{ll}}
	\newcommand{\eeal}{\end{array} \eeq}
\newcommand{\beas}{\begin{eqnarray*}}
	\newcommand{\eeas}{\end{eqnarray*}}
\newcommand{\ba}{\begin{array}}
	\newcommand{\ea}{\end{array}}
\newcommand{\bit}{\begin{itemize}}
	\newcommand{\eit}{\end{itemize}}
\newcommand{\ben}{\begin{enumerate}}
	\newcommand{\een}{\end{enumerate}}

\newcommand{\calA}{{\cal A}}
\newcommand{\calB}{{\cal B}}
\newcommand{\calC}{{\cal C}}
\newcommand{\calD}{{\cal D}}

\newcommand{\calF}{{\cal F}}

\newcommand{\calH}{{\cal H}}
\newcommand{\calI}{{\cal I}}
\newcommand{\calJ}{{\cal J}}

\newcommand{\calM}{{\cal M}}
\newcommand{\calN}{{\cal N}}
\newcommand{\calO}{{\cal O}}

\newcommand{\calS}{{\cal S}}
\newcommand{\calT}{{\cal T}}

\newcommand{\setA}{\textsf{A}}

\newcommand{\setG}{\textsf{G}}
\newcommand{\setH}{\textsf{H}}

\newcommand{\setW}{\textsf{W}}
\newcommand{\setX}{\textsf{X}}

\newcommand{\eg}{\emph{e.g.,}\xspace}
\newcommand{\ie}{\emph{i.e.,}\xspace}
\newcommand{\myParagraph}[1]{{\bf #1.}\xspace}

\newcommand{\M}[1]{{\bm #1}} 
\renewcommand{\boldsymbol}[1]{{\bm #1}}

\newcommand{\hide}[1]{}

\newcommand{\hiddenText}{{\color{gray} hidden text.}}
\newcommand{\hideWithText}[1]{\hiddenText}

\DeclareMathOperator*{\argmax}{arg\,max}

\newcommand{\zero}{{\mathbf 0}}

\newcommand{\MI}{\M{I}}

\newcommand{\MX}{\M{X}}

\newcommand{\MW}{\M{W}}

\newcommand{\vh}{\boldsymbol{h}}

\newcommand{\vxx}{\boldsymbol{x}}
\newcommand{\vy}{\boldsymbol{y}}
\newcommand{\vw}{\boldsymbol{w}}

\newcommand{\blue}[1]{{\color{blue}#1}}
\newcommand{\green}[1]{{\color{green}#1}}
\newcommand{\red}[1]{{\color{red}#1}}

\newcommand{\linkToPdf}[1]{\href{#1}{\blue{(pdf)}}}
\newcommand{\linkToPpt}[1]{\href{#1}{\blue{(ppt)}}}
\newcommand{\linkToCode}[1]{\href{#1}{\blue{(code)}}}
\newcommand{\linkToWeb}[1]{\href{#1}{\blue{(web)}}}
\newcommand{\linkToVideo}[1]{\href{#1}{\blue{(video)}}}
\newcommand{\linkToMedia}[1]{\href{#1}{\blue{(media)}}}
\newcommand{\award}[1]{\xspace} 

\newcommand{\vz}{\boldsymbol{z}}

\graphicspath{{./figures/}}

\IEEEoverridecommandlockouts

\begin{document}
	
	\title{Neural Trees for Learning on Graphs}
	
	\author{Rajat Talak, Siyi Hu, Lisa Peng, and Luca Carlone
		\thanks{The authors are with the Laboratory of Information and Decision Systems (LIDS), Massachusetts Institute of Technology, Cambridge, MA 02139, USA. {\tt \{talak, siyi, lisapeng, lcarlone\}@mit.edu}}
	}

	\IEEEaftertitletext{\vspace{-0.7\baselineskip}}
	
	\maketitle
	\begin{tikzpicture}[overlay, remember picture]
	\path (current page.north east) ++(-4.0,-0.4) node[below left] {
		This paper has been accepted for publication in NeurIPS 2021. Please cite the paper as:
	};
	\end{tikzpicture}
	\begin{tikzpicture}[overlay, remember picture]
	\path (current page.north east) ++(-2.9,-0.9) node[below left] {
		R. Talak, S. Hu, L. Peng, and L. Carlone, ``Neural Trees for Learning on Graphs’’, \emph{NeurIPS}, 2021.
	};
	\end{tikzpicture}
	
	\newcommand{\Htree}{H-tree\xspace}
	\newcommand{\Htrees}{H-trees\xspace}
	\renewcommand{\setA}{\calA}
	\newcommand{\qed}{{\hfill $\square$}}
	\newcommand{\NTlong}{neural tree\xspace}
	\newcommand{\NTslong}{neural trees\xspace}
	\newcommand{\NT}{NT\xspace}
	\newcommand{\NTs}{NTs\xspace}
	\newcommand{\JNTlong}{neural tree\xspace}
	\newcommand{\JNT}{neural tree\xspace}
	\newcommand{\JNTs}{neural trees\xspace}
	\newcommand{\Graph}{\ensuremath{\mathcal{G}}\xspace}
	\newcommand{\Nodes}{\ensuremath{\mathcal{V}}}
	\newcommand{\Edges}{\ensuremath{\mathcal{E}}}
	\newcommand{\JTH}[1]{\ensuremath{\mathcal{J}_{#1}}}
	\newcommand{\Vgets}{\ensuremath{\leftarrow}\xspace}
	\newcommand{\AGG}{\text{AGG}}
	\newcommand{\COMB}{\text{COMB}}
	\newcommand{\FUSE}{\text{FUSE}}
	\newcommand{\SetOfLabels}{\ensuremath{\mathcal{L}}}
	\newcommand{\CliqueSetOf}[1]{\ensuremath{\mathcal{C}\left(#1\right)}}
	\newcommand{\NumIterations}{\ensuremath{T}} 
	\newcommand{\DataSet}{\ensuremath{\mathcal{D}}}
	\newcommand{\DataSize}{\ensuremath{D}}
	\newcommand{\LabelClass}{\ensuremath{\mathcal{L}}}
	\newcommand{\nbr}[2]{\ensuremath{\calN_{#2}\left(#1\right)}}
	\newcommand{\READ}{\text{READ}}
	\newcommand{\EgoGraph}[2]{\ensuremath{\Graph_{#1,#2}}}
	\newcommand{\treewidth}[1]{\ensuremath{\text{tw}\left[ #1 \right]}}
	\newcommand{\featureSubset}{\MX}
	\newcommand{\nodeSubset}{\calA}
	\newcommand{\bag}{\calB}
	\newcommand{\Tree}{\ensuremath{\calT}}
	\newcommand{\Root}{R\xspace}
	
	\newcommand{\dJTH}[1]{\ensuremath{\vec{\JTH{#1}}}}
	\newcommand{\ReLU}[1]{\ensuremath{\texttt{ReLU}\left(#1\right)}}
	\newcommand{\inner}[2]{\ensuremath{\langle #1, #2\rangle}}
	
	\newcommand{\treeDecomposition}{tree decomposition\xspace}
	\newcommand{\treeDecompositions}{tree decompositions\xspace}
	
	\newcommand{\suppMatt}{supplementary material\xspace}
	
	\newcommand{\edited}[1]{\red{#1}}
	
	\renewcommand{\setX}{{\mathbb X}} 
	
	\newcommand{\homnum}[1]{\ensuremath{\text{hom}\left( #1 \right)}}
	\newcommand{\homnumX}[1]{\ensuremath{\overline{\text{hom}}\left( #1 \right)}}
	
	\newcommand{\Homnum}[2]{\ensuremath{\text{HOM}_{#2}\left( #1 \right)}}
	\newcommand{\HomnumX}[2]{\ensuremath{\overline{\text{HOM}}_{#2}\left( #1 \right)}}

	\renewcommand{\omit}[1]{}
	\newcommand{\add}[1]{\green{#1}}
	
	\newcommand{\modify}[1]{\textcolor[rgb]{0.50,0.00,0.00}{#1}}

	\begin{abstract}
		Graph Neural Networks (GNNs) have emerged as a flexible and powerful approach for learning over graphs.
		Despite this success, existing GNNs are constrained by their local message-passing architecture and are provably limited in their expressive power.
		In this work, we propose a new GNN architecture -- the \emph{Neural Tree}.
		The neural tree architecture does not perform message passing on the input graph, but on a tree-structured graph, called the \emph{H-tree}, that is constructed from the input graph.
		Nodes in the \Htree correspond to subgraphs in the input graph, and they
		are reorganized in a hierarchical manner such that the parent of a node in the \Htree always corresponds to a larger subgraph in the input graph.
		We show that the neural tree architecture can approximate any smooth probability distribution function over an 
		undirected graph\omit{ as well as emulate the junction tree algorithm}.
		We also prove that the number of parameters needed to achieve an $\epsilon$-approximation of the distribution function is  exponential in the treewidth of the input graph, but linear in its size.
		We prove that any continuous \Graph-invariant/equivariant function can be approximated by a 
		nonlinear combination of
		such probability distribution functions over \Graph.
		We apply the neural tree to semi-supervised node classification in 3D scene graphs, and show that these theoretical properties translate into significant gains in prediction accuracy, over the more traditional GNN architectures.
		We also show the applicability of the neural tree architecture to citation networks with large treewidth, by using a graph sub-sampling technique.
	\end{abstract}


	\section{Introduction}
	\label{sec:intro}
	
	%
	Graph-structured learning problems arise in several disciplines,
	including biology (\eg molecule classification~\cite{Duvenaud15neurips-gnnMolecule}),
	computer vision (\eg
	action recognition~\cite{Guo18eccv},
	image classification~\cite{Garcia18iclr-fewshot}, shape and pose estimation~\cite{Kolotouros19cvpr-shapeRec}),
	computer graphics (\eg  mesh and point cloud classification and segmentation~\cite{Hanocka19acm-meshcnn,Li19iccv-deepgcns,Milano20neurips-PDMeshNet}),
	and social networks (\eg fake news detection~\cite{Monti19iclrws}), among others~\cite{Bronstein17spm-geometricDL}.
	In this landscape, \emph{Graph Neural Networks} (GNN) have gained popularity as a
	flexible and effective approach for regression and classification over graphs.
	
	Despite this growing research interest,
	recent work has pointed out several limitations of existing GNN architectures~\cite{Xu19iclr-gin, Morris2019j-AAAI-HigherOrderGNN,Maron2019c-NeurIPS-ProvablyPowerfulGNN, Giorgos2020a-arXiv-ImprovingGNN-SubgraphIso}.
	%
	Local message passing GNNs are no more expressive than the Weisfeiler-Lehman (WL) graph isomorphism test~\cite{Xu19iclr-gin}, neither can they serve as universal approximators to all \Graph-invariant (equivariant) functions, \ie functions defined over a graph \Graph that remain unchanged by (or commute with) node permutation. The work~\cite{Chen19neurips} proves an equivalence between the ability to do graph isomorphism testing and the ability to approximate any \Graph-invariant function.
	
	Various GNN architectures have been proposed, that go beyond local message passing or use tensor representations, in order to improve expressivity. Graph isomorphism testing, \Graph-invariant/equivariant function approximation, and the generalized $k$-order WL ($k$-WL) tests have served as end objectives and guided recent progress of this inquiry.
	%
	%
	For example, $k$-order linear GNN~\cite{Maron19iclr} and $k$-order folklore GNN~\cite{Maron2019c-NeurIPS-ProvablyPowerfulGNN} have expressive powers equivalent to $k$-WL and $(k+1)$-WL test, respectively~\cite{Azizian21iclr-gnn-expressivity}. While these architectures can theoretically approximate any \Graph-invariant function (as $k \rightarrow \infty$), they use $k$-order tensors for representations, rendering them impractical for any $k > 3$.
	
	There is a need for a new way to look at constructing GNN architectures.
	With better end objectives to guide theoretical progress. Such an attempt can
	result in new and expressive GNNs that are provably tractable -- if not in general, at least in reasonably constrained settings.

	
	A GNN, by its very definition, operates on graph structured data.
	The graph structure of the data determines inter-dependency between nodes and their features. Probabilistic graphical models present a reasonable and well-established way of articulating and working with such inter-dependencies in the data. Prior to the advent of neural networks, inference algorithms on such graphical models were successfully applied to many real-world problems. Therefore, we pose that a GNN architecture operating on a graph should have at least the expressive power of  a probabilistic graphical model, \ie it should be able to approximate any distribution defined by a probabilistic graphical model.
	
	This is not a trivial requirement as exact inference (akin to learning the distribution or its marginals) on a probabilistic graphical model, without any structural constraints on the input graph, is known to be an NP-hard problem~\cite{Cooper90ai}. Even approximate inference on a probabilistic graphical model is known to be NP-hard
	in general~\cite{Roth96ai}. 
	%
	%
	%
	A common trick to perform exact inference, consists in constructing a \emph{junction tree} for an input graph and performing message passing on the junction tree instead. In the junction tree, each node corresponds to a subset of nodes of the input graph.
	The junction tree algorithm remains tractable for graphs with bounded treewidth, while \cite{venkat12uai} shows that treewidth is the only structural parameter, bounding which, allows for tractable inference on graphical models.
	
	\myParagraph{Contribution}
	We first define the notion of \Graph-compatible function and argue that approximating \Graph-compatible functions is equivalent to approximating any probability distribution on a probabilistic graphical model (Section~\ref{sec:compatible-functions}); we also show that \Graph-invariant/equivariant functions considered
	in related work can be approximated using a nonlinear combination of \Graph-compatible functions.
	
	
	We then propose a novel GNN architecture -- the \emph{Neural Tree} -- that can approximate any \Graph-compatible function (Section~\ref{sec:architecture}). Neural trees do not perform message passing on the input graph, but on a tree-structured graph,
	called the \emph{\Htree}, that is constructed from the input graph.
	Each node in the \Htree corresponds to a subgraph of the input graph. These subgraphs are arranged hierarchically in the \Htree such that the parent of a node in the \Htree always corresponds to a larger subgraph in the input graph.
	The leaf nodes in the \Htree correspond to singleton subsets (\ie individual nodes) of the input graph.
	The \Htree is constructed by recursively computing \treeDecompositions of the input graph and its subgraphs, and attaching them to one another to form a hierarchy.
	Neural message passing on the \Htree generates representations for all the nodes and important subgraphs of the input graph.
	
	We next prove that the neural tree architecture can approximate any smooth \Graph-compatible function defined over a given undirected graph (Section~\ref{sec:approximation_results}).
	We also bound the number of parameters required by a neural tree architecture to obtain an $\epsilon$-approximation of an arbitrary (smooth) \Graph-compatible function.
	We show that the number of parameters increases exponentially in the treewidth of the input graph, but only linearly in the input graphs size. 
	Thus, for graphs with bounded treewidth, the neural tree can tractably approximate any smooth distribution function.
	
	\omit{The proposed neural tree architecture has a message passing structure that can emulate a junction tree algorithm. This is unlike the standard, local message passing GNN architectures.
		\cite{Xu20iclr-GNN-algoAlign} suggests that such algorithmic alignments of neural architectures lead to lower sample complexity and better generalizability. Thus, for the task of inference on graphs, the neural tree architecture may also provide for better generalizablility.}
	
	We apply the neural tree architecture for
	semi-supervised node classification in 3D scene graphs and citation networks (Section~\ref{sec:experiments}).
	Our experiments on 3D scene graphs demonstrate that neural trees outperform standard, local message passing GNNs, by a large margin.
	Citation networks on the other hand, typically have large treewidth;
	therefore we make use of a recently proposed bounded treewidth
	graph sub-sampling algorithm~\cite{Yoo20wsdm-graphSubsampling}, that sub-samples the input graph (\ie removes edges) to reduce its treewidth to a specified number. We show that applying the neural tree architecture in conjunction with such sub-sampling algorithm makes our architecture scalable to large graphs while
	still preserving its advantage over traditional architectures.
	%
	Our code is publically available at \url{https://github.com/MIT-SPARK/neural_tree}

	\section{Related Work}
	\label{sec:related_work}

	\myParagraph{Expressive Power of Graph Neural Networks}
	%
	%
	Since the seminal works~\cite{Gori2005gnn, Scarselli08tnn-gnn}, various GNN architectures have been proposed including Graph Convolutional Networks (GCN)~\cite{Kipf17iclr-gcn, Henaff15-deep, Defferrard16nips-ChebyNets, Kipf17iclr-gcn, Bronstein17spm-geometricDL}, Message Passing Neural Networks (MPNN)~\cite{Gilmer17icml-mpnns}, GraphSAGE~\cite{Hamilton17nips-GraphSage}, Graph Attention Networks (GAT)~\cite{Velickovic18iclr-GAT, Lee19tkdd-gat-survey, Busbridge2019a-arXiv-Rel-GraphAttentionNetworks}, message passing GNN~\cite{Gilmer17icml-mpnns}. \omit{Their successful applications to various problems
		have led to a renewed interest towards GNNs.}
	%
	%
	Limited expressive power of these standard GNNs has been a major concern.
	For instance, it is known that local message passing GNNs can neither distinguish between non-isomorphic graphs (provably worse than the 1-Weisfeiler-Lehman (WL) test)~\cite{Xu19iclr-gin, Morris2019j-AAAI-HigherOrderGNN}, nor can they compute even simple graph properties~\cite{Garg2020c-ICML-GenAndRep-GNN}. \omit{such as longest or shortest cycle, diameter, or clique information~\cite{Garg2020c-ICML-GenAndRep-GNN}. } 
	
	Many GNN architectures have been proposed to overcome this expressivity bottleneck. 
	Graph substructure network is proposed in
	\cite{Giorgos2020a-arXiv-ImprovingGNN-SubgraphIso} and is shown to be more powerful than the 1-WL test.
	$k$-order GNNs, in which message passing is performed among a subset of nodes in the input graph, is shown to have expressive power equivalent to the generalized $k$-WL test~\cite{Morris2019j-AAAI-HigherOrderGNN, Maron2019c-NeurIPS-ProvablyPowerfulGNN}.
	%
	It is generally understood that to improve the expressivity of GNNs one has to extract features corresponding to important subgraphs, and operate on them.
	A hierarchical architecture that pools a representation vector from a subset of nodes, at each layer, is proposed in~\cite{Ying18nips-DiffPool}, while a hierarchical graph neural network for node clustering is proposed in~\cite{Xing21iccv-clusteringHierarchicalGNN}. A junction-tree based message passing GNN is proposed for molecular graph generation in ~\cite{Jaakkola2018c-ICML-JTVAE-GNN}.
	%

	Graph neural networks have been investigated as function approximators since the beginning. \cite{scarselli09tnn-GnnFunApprox} introduces the notion of \emph{unfolding equivalence} and derives a universal approximation result for graph neural networks. 
	Recent research in developing expressive GNN architectures has been towards approximating graph invariant/equivariant functions~\cite{Maron19icml, Maron19iclr, Maron2019c-NeurIPS-ProvablyPowerfulGNN, Keriven19neurips, Sannai19arxiv-universal}.
	While, invariance and equivariance are desirable properties, the problem of designing GNNs that are universal approximators of \Graph-invariant/equivariant functions has been difficult.
	For instance, the $k$-order GNNs~\cite{Maron19iclr, Maron2019c-NeurIPS-ProvablyPowerfulGNN} can provably
	approximate any graph invariant function, but only as $k \rightarrow \infty$, rendering them impractical~\cite{Azizian21iclr-gnn-expressivity}.
	An equivalence between designing GNN architectures to approximate graph invariant functions and graph isomorphism testing
	is shown in~\cite{Chen19neurips}.
	%
	The generalization power of GNNs has also been investigated in~\cite{scarselli18nn-VC-gnn, Garg2020c-ICML-GenAndRep-GNN, Xu20iclr-GNN-algoAlign, Xu21icml-gnn-extrapolate}.

	\myParagraph{Scene Graphs} Scene graphs are a popular model to abstract information in images or model 3D environments.
	2D scene graphs have been used in
	image retrieval~\cite{Johnson15cvpr}, caption generation~\cite{Karpathy15cvpr-caption, Anderson16eccv-sceneGraphDescription}, visual question answering~\cite{Ren15corr-QA, Krishna16arxiv-visualGenome}, and relationship detection~\cite{Lu16eccv-visualRelations}.
	GNNs are a popular tool for joint object labels and/or relationship inference on scene graphs~\cite{Xu17cvpr-sceneGraph, Li17iccv-sceneGraphGeneration, Yang18eccv-sceneGraph, Zellers18cvpr-sceneGraph}.
	Recently, 
	there has been a growing interest towards \emph{3D} scene graphs, which are constructed from 3D data, such as meshes~\cite{Armeni19iccv-3DsceneGraphs}, point clouds~\cite{Wald20cvpr-semanticSceneGraphs}, or raw sensor data~\cite{Kim19tc-3DsceneGraphs,Rosinol20rss-dynamicSceneGraphs}.
	GNNs have been very recently applied to 3D scene graphs
	for scene layout prediction~\cite{Wald20cvpr-semanticSceneGraphs}
	or object search~\cite{Kurenkov20arxiv-semantic}.

	\section{Problem Statement and Preliminaries}
	\label{sec:preliminaries}
	
	In this section, we state the node classification problem and review standard graph neural networks. 
	
	\myParagraph{Problem} We focus on the standard problem of \emph{semi-supervised node classification}~\cite{Kipf17iclr-gcn}. We are given a graph $\Graph = (\Nodes, \Edges)$ along with node features $\MX = (\vxx_v)_{v \in \Nodes}$; where $\vxx_v$ denotes the node feature of node $v \in \Nodes$. The graph is not necessarily connected. A subset of nodes $\nodeSubset \subset \Nodes$ in $\Graph$ are labeled, \ie {$\{z_{v} \in \LabelClass~|~v\in \nodeSubset \}$} is given;  here $z_v$ denotes the label for node $v$ and $\LabelClass$ the finite set of label classes. We need to design a model to predict the labels of all the unlabeled nodes $v \in \Nodes \backslash \nodeSubset$.
	See Appendix~\ref{sec:app-notations} for the notation used in the paper.

	\myParagraph{Graph Neural Networks (GNN)}
	Various GNN architectures have been successfully applied to solve the node classification problem~\cite{Hamilton17nips-GraphSage, Kipf17iclr-gcn, Velickovic18iclr-GAT, Xu19iclr-gin, Jaakkola2018c-ICML-JTVAE-GNN}.
	Standard GNN architectures construct representation vectors for each node in $\Graph$ by iteratively aggregating representation vectors of its neighboring nodes.
	%
	At iteration $t$, the representation vector of node $v$ is generated as follows:
	\begin{equation}
		\label{eq:GNN_AGG}
		h^{t}_{v} = \AGG_{t}\left( h^{t-1}_{v}, \{ \left( h^{t-1}_{u}, \kappa_{u,v} \right)~|~u \in \nbr{v}{\Graph} \}\right),
	\end{equation}
	with $h^{0}_{v} \triangleq \vxx_{v}, \, \forall v \in \Nodes$; where $\nbr{v}{\Graph}$ denotes the set of neighbors of node $v$ in graph \Graph~and the aggregation function $\AGG_{t}$ can depend on the trainable edge parameters $\kappa_{u,v}$.
	%
	%
	This process of sharing and aggregating representation vectors among neighboring nodes in $\Graph$ is often called \emph{message passing}.
	This procedure runs for a fixed number of iterations $\NumIterations$. The node labels are then generated from the representation vectors at the final iteration $\NumIterations$. 
	Node labels are extracted as
	\begin{equation}
		\label{eq:GNN_READ}
		y_{v} = \READ(h^{T}_{v}),
	\end{equation}
	for all $v \in \Nodes$. 
	%
	%
	The functions $\AGG_t$ and $\READ$ are modeled as single or multi-layer perceptrons. 

	\section{Graph Compatible Functions}
	\label{sec:compatible-functions}
	
	We start by defining a class of \Graph-compatible functions.
	\Graph-compatible functions allow us to establish connections with probabilistic graphical models and
	the \Graph-invaraint/equivariant functions.
	
	
	\begin{definition}[\Graph-compatible functions]
		\label{def:compatible_fun}
		We say that a function $f:(\times_{v \in \Nodes} \setX_v, \Graph) \rightarrow \mathbb{\Root}$ is compatible with graph $\Graph$ or \Graph-compatible if it can be factorized as
		\begin{equation}\label{eq:compatible}
			f(\MX) = \textstyle\sum_{C \in \CliqueSetOf{\Graph}}\theta_{C}(\vxx_C),
		\end{equation}
		where $\CliqueSetOf{\Graph}$ denotes the collection of all maximal cliques in $\Graph$ and $\theta_{C}$ is some function that maps $\times_{v \in C} \setX_v$ (the set of node features in the clique $C$) to a real number.
	\end{definition}
	
	Compatible functions arise in probabilistic graphical models; for instance, the logarithm of a joint probability distribution is a compatible function (see Appendix~\ref{sec:app-inferenceInGraphicalModels} for more examples on how such functions arise in inference on graphical models).

	\subsection{Relation with Invariant/Equivariant Functions}
	A graph invariant function requires that the function output remains invariant to node permutation,  whereas a graph equivariant function outputs a vector (or a tensor in general) which is required to commute with any permutation applied to the input graph nodes.
	While graph invariance is a desirable property for graph classification problems, graph equivariance is desirable in node classification problems.
	
	%
	
	%
	%
	%
	We now show that any continuous \Graph-invariant or \Graph-equivariant function can be written as a finite sum of \Graph-compatible functions, each composed with a specific nonlinear
	function.  The precise definitions of \Graph-invariant and \Graph-equivariant functions are given in Appendix~\ref{sec:app-invariant-equivariant}.
	\begin{theorem}[Invariance/Equivariance]
		\label{thm:invariant-equivariant}
		The following statements hold true.
		\begin{enumerate}
			\item For any continuous \Graph-invariant function $h:\setX \rightarrow \mathbb{R}$ and an $\epsilon > 0$ there exists an integer $M \geq 1$ and a collection of $M$ continuous \Graph-compatible functions $\{ f^{i} \}_{i=1}^{M}$ such that
			\begin{equation}
				\sup_{\MX \in \setX}~~\left| h(\MX) - \sum_{i=1}^{M}\phi\left( f^{i}(\MX)\right)\right| < \epsilon,
			\end{equation}
			where $\phi : \mathbb{R} \rightarrow \mathbb{R}$ is some function.
			\item For any continuous \Graph-equivariant function $h:\setX \rightarrow \mathbb{R}^n$ and an $\epsilon > 0$ there exists a set of integers $M_l \geq 1$, for $l \in [n]$, and \Graph-compatible functions $\{ f^{l,i}\}_{i=1}^{M_l}$ such that
			\begin{equation}
				\sup_{\MX \in \setX}~~\left| h_l(\MX) - \sum_{i=1}^{M_l}\phi\left( f^{l,i}(\MX)\right)\right| < \epsilon,
			\end{equation}
			for all $l \in [n]$, where $h_l(\MX) \in \mathbb{R}$ denotes the $l$th component of $h$ and $\phi : \mathbb{R} \rightarrow \mathbb{R}$ is some function.
		\end{enumerate}
	\end{theorem}
	\begin{IEEEproof}
		See Appendix~\ref{sec:app-invariant-equivariant}.
	\end{IEEEproof}
	This result shows that a GNN architecture that can approximate any \Graph-compatible function will also be able to approximate graph invariant and equivariant function. 
	%


	In the next section, we describe the neural tree architecture, which can approximate any (smooth) \Graph-compatible function.
	%

	\section{Neural Tree Architecture}
	\label{sec:architecture}

	The key idea behind the neural trees architecture is to construct a tree-structured graph from the input graph 
	and perform message passing on the resulting tree instead of the input graph.
	This tree-structured graph is such that every node in it represents either a node or a subset of nodes in the input graph.
	Trees are known to be more amenable for message passing~\cite{Jordan02book, Koller09book} and indeed
	the proposed architecture enables the derivation of strong approximation results,
	which we present in Section~\ref{sec:approximation_results}. 



	In the following, we first review the notion of \treeDecomposition (Section~\ref{sec:treeDecomposition}). We then show how to construct a \Htree for a graph, by successively applying \treeDecomposition on a given graph $\Graph$ and its subgraphs (Section~\ref{sec:jth_gen}). Finally, we discuss the proposed neural tree architecture for node classification, which performs neural message passing on the \Htree (Section~\ref{sec:architecture-message-passing}).

	In Section~\ref{sec:approximation_results}, we show that the tree structure enables the derivation of strong approximation results by which a neural tree can approximate any (smooth) \Graph-compatible function.
	
	\subsection{Tree Decomposition}
	\label{sec:treeDecomposition}
	
	For a graph $\Graph$, a \emph{\treeDecomposition} is a tuple $(\Tree, \bag)$ where $\Tree$ is a tree graph and $\bag = \{ B_{\tau} \}_{\tau \in \Nodes(\Tree)}$ is a family of \emph{bags}, where $B_{\tau} \subset \Nodes(\Graph)$ for every tree node $\tau \in \Nodes(\Tree)$, such that the tuple $(\Tree, \bag)$ satisfies the following two properties:
	
	(1) \emph{Connectedness:} for every graph node $v \in \Nodes(\Graph)$,
	the subgraph of $\Tree$ induced by tree nodes $\tau$ whose bag contains node $v$, is connected, \ie
	$\Tree_v \triangleq \Tree\left[\{ \tau \in \Nodes(\Tree)~|~v \in B_{\tau}\}\right]$
	is a connected subgraph of $\Tree$ for every $v \in \Nodes(\Graph)$.
	
	(2) \emph{Covering:} for every edge $\{u, v\}$ in $\Graph$ there exists a node $\tau \in \Nodes(\Tree)$ such that $u, v \in B_{\tau}$.
	
	The simplest \treeDecomposition of any graph $\Graph$ is a tree with a single node, whose bag contains all the nodes in $\Graph$. However, in practical applications,
	it is desirable to obtain decompositions where the size of the largest bag is small.
	This is captured by the notion of \emph{treewidth}.
	The treewidth of a \treeDecomposition $(\Tree, \bag)$ is defined as the  size of the largest bag minus one:
	\begin{equation}
		\label{eq:def_treewidth}
		\textstyle{\treewidth{(\Tree, \bag)} \triangleq \max_{\tau \in \Nodes(\Tree)} |B_{\tau}| - 1.}
	\end{equation}
	The treewidth of a graph $\Graph$ is defined as the minimum treewidth that can be achieved among all \treeDecompositions of $\Graph$. While
	finding a \treeDecomposition with minimum treewidth is NP-hard, 
	many algorithms exist that generate \treeDecompositions with small enough treewidth~\cite{Becker1996c-UAI-FastApproxOptJT, Bodlaender2006c-Chapter-TreewidthComputation, Thomas2009j-JCGS-EnumerateJT, Bodlaender2010j-IC-TreewidthComputationI, Bodlaender2010j-IC-TreewidthComputationII}.
	
	We use $(\Tree, \bag) = \texttt{tree-decomposition}(\Graph)$  to denote a generic tree decomposition of a graph \Graph.
	One of the most popular \treeDecompositions is the junction tree decomposition, which was introduced in~\cite{Jensen1994c-UAI-OptimalJT}. We denote it by $(\Tree, \bag) = \texttt{junction-tree}(\Graph)$ and describe it's construction in Appendix~\ref{sec:app-junctionTree}
	for completeness.
	\subsection{\Htree}
	\label{sec:jth_gen}
	\begin{figure}
		\centering
		\includegraphics[width=\linewidth]{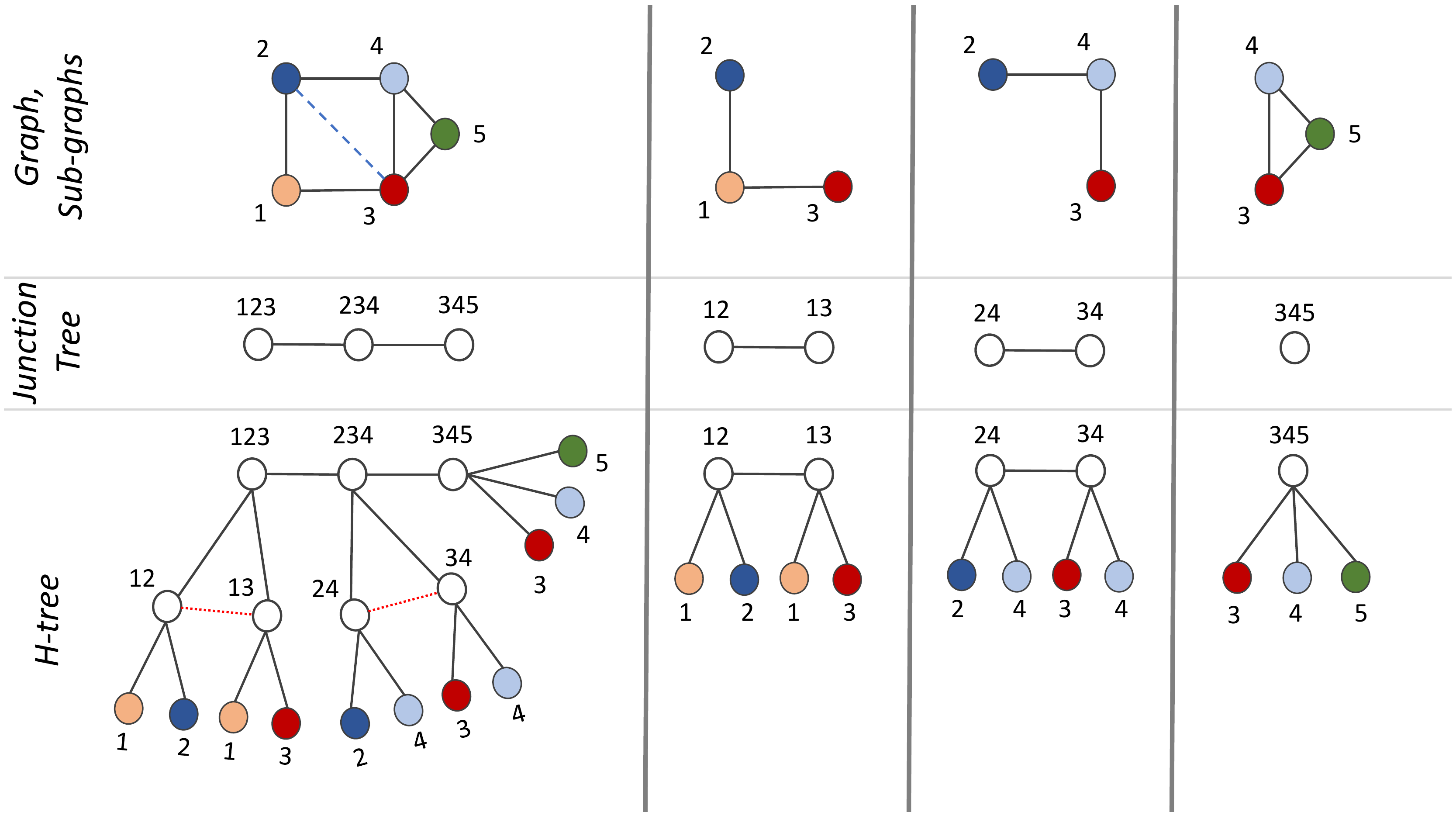}
		\caption{Generating \Htrees for graph $\Graph$ and its subgraphs.}
		\label{fig:jth-illustrate}
	\end{figure}
	We first define \Htree for a complete graph. Let $\calS_{n}$ denote a star graph with $n$ leaf nodes and one root. 
	\begin{definition}[Complete graph]
		\label{def:completeGraph_jth}
		For a complete graph \Graph with $n$ nodes, the \Htree is a star graph, \ie $\JTH{\Graph} = \calS_{n}$, where the root node (in \JTH{\Graph}) represents the single maximal clique in $\Graph$ and each of the leaf nodes in $\calS_{n}$ corresponds to a node in $\Graph$.
	\end{definition}
	
	The \Htree for a complete graph of three nodes is shown in Fig.~\ref{fig:jth-illustrate}, rightmost column. In it, the unique clique in the graph, which contains nodes $\{3, 4, 5\}$, is labeled as $C = (345)$. 
	For the sake of clarity, we always enlist the set of root nodes when defining an \Htree. Therefore, an \Htree of a graph \Graph is given by a tuple $(\JTH{\Graph}, \Root)$, where
	$\JTH{\Graph}$ is a tree graph and $\Root$ is the set of \emph{root} nodes.

	%
	%
	The \Htree is computed by recursively applying tree decomposition on the input graph and the subgraphs obtained in tree decomposition. For instance, if $(\Tree, \calB)$ is a tree decomposition of the input graph \Graph, then we recursively apply tree decomposition to each subgraph $\Graph[B]$ (of \Graph) for each $B \in \calB$. 
	The final \Htree \JTH{\Graph} is computed by connecting all the obtained tree decomposition as a hierarchy. The set of root nodes $R$ are the nodes in \JTH{\Graph} corresponding to the original tree decomposition $(\Tree, \calB)$ of the graph. This process is illustrated in Figure~\ref{fig:jth-illustrate} and the algorithm described in Algorithm~\ref{algo:jth}.

	We now describe the algorithm in more detail. Algorithm~\ref{algo:jth} takes  an undirected graph $\Graph$ and outputs a \Htree $\JTH{\Graph}$ with a set of root nodes $R$. Let $(\Tree, \calB)$ denote a \treeDecomposition of graph $\Graph$ (line~\ref{algo:line:treeB}). The \Htree $\calJ$ is initialized to $\calJ = \Tree$ (line~\ref{algo:line:JeqT}) and the set of root nodes equals the root nodes of this tree, namely $R = \Nodes(\Tree)$ (line~\ref{algo:line:root}).
	For $B \in \calB$, {let $\tau(B)$ denote the node corresponding to bag $B$ in $\calJ$}.
	Then for each bag $B \in \calB$ we construct a \Htree of the induced subgraph $\Graph[B]$ (lines~\ref{algo:line:beginJTH}-\ref{algo:line:endJTH}).

	If $\Graph[B]$ is not complete, we attach its root nodes to $\tau(B)$ (lines~\ref{algo:line:beginElse}-\ref{algo:line:endElse}). Specifically, if $(\calJ', R')$ denotes the \Htree for the induced subgraph $\Graph[B]$, then we attach the graph $\calJ'$ to $\calJ$ by linking all root nodes of $\calJ'$, namely $R'$, to the node $\tau(B)$ (lines~\ref{algo:line:beginLinkRoots}-\ref{algo:line:endLinkRoots}). To avoid cycles, we also remove edges between the root nodes $R'$ in $\calJ'$(line~\ref{algo:line:removeRootEdges}).
	
	If the induced subgraph $\Graph[B]$ is complete, then from Definition~\ref{def:completeGraph_jth} we know that its \Htree is a star graph with a single clique node, call it $C$. In this case, we attach the star graph to $\tau(B)$ by merging two nodes -- $C$ and $\tau(B)$ -- into one. This avoids an unnecessary edge $(\tau(B), C)$ in the \Htree.

	\textbf{Example.} Figure~\ref{fig:jth-illustrate} shows the construction of a \Htree for a graph with $5$ nodes and $6$ edges. Here, we have used the \texttt{junction-tree} algorithm to perform \treeDecomposition. The first column shows the graph $\Graph$ and its junction trees, which has three nodes corresponding to the three cliques in the chordal graph $\Graph_{c}$ (which in this case consists in adding the dashed blue line in Figure~\ref{fig:jth-illustrate}; see Appendix~\ref{sec:app-junctionTree}
	for details).\footnote{The chordal graph $\Graph_{c}$ is used in the junction tree construction and is obtained from $\Graph$ after graph triangulation, which in this case consists in adding the dashed blue line in Figure~\ref{fig:jth-illustrate}.}
	The remaining columns show the three subgraphs of $\Graph$ corresponding to each of the three maximal cliques in $\Graph_{c}$, along with their junction trees and \Htrees. The \Htree of each of these subgraphs is then attached to the junction tree of $\Graph$ to get the required \Htree for $\Graph$. The \Htree for graph $\Graph$ is shown in the last row of the first column in Figure~\ref{fig:jth-illustrate}. Also illustrated are the two edges deleted (in red) when merging the two \Htrees of the subgraphs to the junction tree of $\Graph$.

	\begin{algorithm}[h]
		\SetAlgoLined
		\SetKwInOut{Input}{input}\SetKwInOut{Output}{output}
		\Input{Graph $\Graph$}
		\Output{\Htree~$(\JTH{\Graph}, \Root)$\hspace{-5mm}} 
		%
		%
		$(\Tree, \calB) \Vgets \texttt{tree-decomposition}(\Graph)$ \label{algo:line:treeB}\\
		$\calJ$ \Vgets \Tree \label{algo:line:JeqT}\\
		$\Root$ \Vgets \Nodes(\Tree) \label{algo:line:root}\\
		\For{\emph{each bag} $B$ \emph{in} $\calB$}{ \label{algo:line:beginJTH}
			
			\eIf{$\Graph[B]$ \emph{is a complete graph}}{
				Update $\calJ$:\\
				$\Nodes(\calJ) \Vgets \Nodes(\calJ)\cup B$ \\
				$\Edges(\calJ) \Vgets \Edges \cup \{ \{ \tau(B), b \} \}_{b \in B}$ \\
				%
			}
			{\label{algo:line:beginElse}
				$(\calJ', \Root') \Vgets \texttt{H-tree}(\Graph[B])$ \\
				Update $\calJ$: \\
				$\Edges(\calJ') \Vgets \Edges(\calJ')\backslash \Edges(\calJ'[\Root'])$ \label{algo:line:removeRootEdges}\\
				$\calJ \Vgets \calJ \cup \calJ'$ \label{algo:line:beginLinkRoots}\\
				$\Edges(J) \Vgets \Edges(J)\cup \{ \{\tau(B), r\} \}_{r \in \Root'}$ \label{algo:line:endLinkRoots}\\
			}\label{algo:line:endElse}
			$\JTH{G} \Vgets \calJ$ \\
			return (\JTH{G}, \Root) \label{algo:line:endJTH}
			\BlankLine
			
		}
		\caption{\texttt{H-tree}}
		\label{algo:jth}
	\end{algorithm} 
	
	\begin{remark}[Leaves and features]
		Each node in the \Htree (of a graph $\Graph$) corresponds to a subset of nodes in graph $\Graph$. Every leaf node $l$ in $\JTH{\Graph}$ corresponds to exactly one node $v$ in $\Graph$. We denote this node by $\kappa(l)$ for every leaf node $l$ of $\JTH{\Graph}$.
		In the construction of the \Htree, we also assign the node input feature $\vxx_{v}$ to every node $l$ in $\JTH{\Graph}$ for which $\kappa(l) = v$. {Note that multiple leaf nodes may correspond to a single node $v$ in the graph $\Graph$, \ie we can have $\kappa(l) = v$ for many leaf nodes $l$ in $\JTH{\Graph}$.}
		Fig.~\ref{fig:jth-illustrate} illustrates the input node features by node coloring. \omit{We show the input node features being replicated on {(multiple)} leaf nodes of the \Htree.
			The input node features for other nodes in $\JTH{\Graph}$ is set to $\zero$.}
	\end{remark}

	\subsection{Message Passing on \Htree}
	\label{sec:architecture-message-passing}
	
	\omit{
		\begin{wrapfigure}{r}{0.4\textwidth}
			\vspace{-0.5cm}
			\centering
			\includegraphics[width=0.98\linewidth]{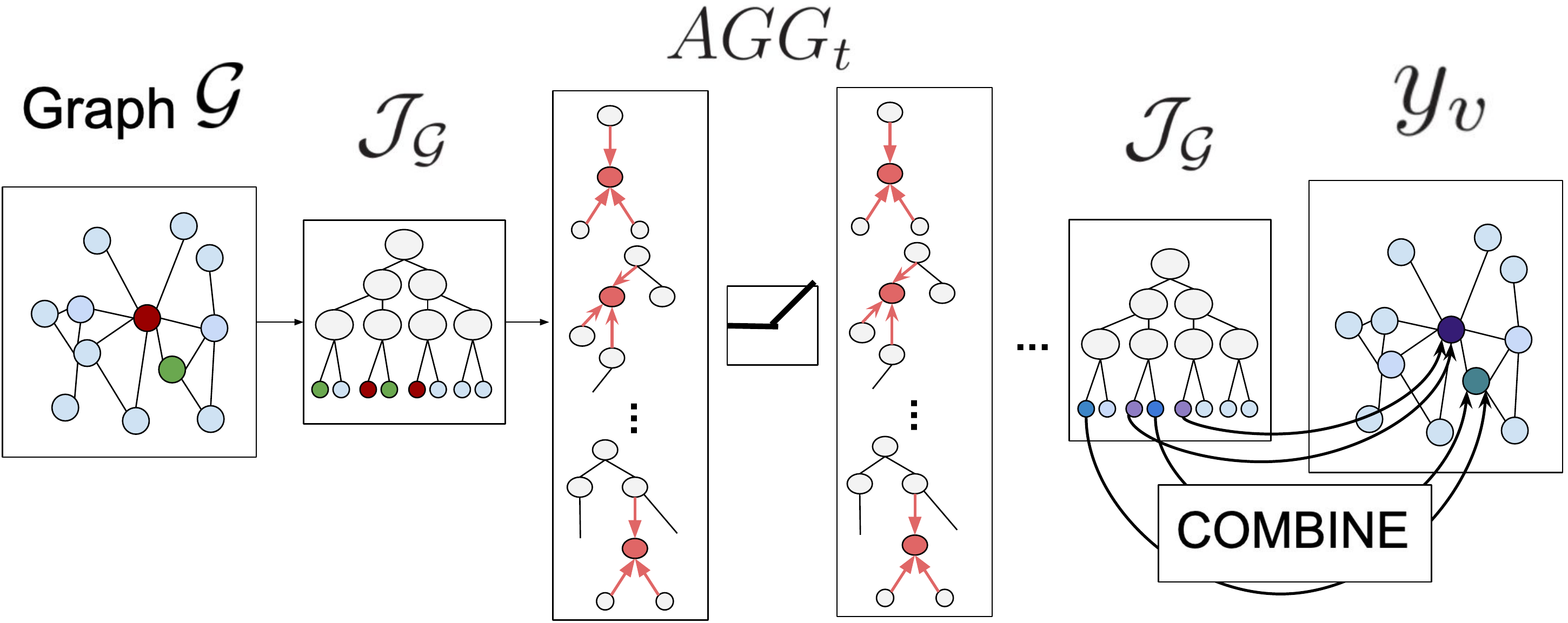}
			\caption{Neural tree architecture for node classification.}\label{fig:jnt-node-classif-architecture-illlustrate}
		\end{wrapfigure}
	}
	Given a graph $\Graph$ with input node features, we construct a \Htree $(\JTH{\Graph}, R)$ and perform message passing on $\JTH{\Graph}$. We call this the \emph{neural tree} architecture. Representation vectors are generated for each node in the \Htree $\JTH{\Graph}$ by aggregating representation vectors of neighboring nodes in $\JTH{\Graph}$.
	The message passing starts with $h^{0}_{l} = \vxx_{\kappa(l)}$ for all leaf nodes $l$ in $\JTH{\Graph}$ and $h^{0}_{u} = \zero$ for non-leaf nodes $u$ in $\JTH{\Graph}$. These representation vectors are then updated as
	\begin{equation}\label{eq:agg}
		\vh^{t}_{u} = \AGG_t\left(\vh^{t-1}_{u}, \{ \left( \vh^{t-1}_{w}, \kappa_{w, u} \right)~|~w \in \nbr{u}{\JTH{\Graph}} \}\right),
	\end{equation}
	for each iteration $t \in \{1, 2, \ldots \NumIterations \}$. The aggregation function $\AGG_{t}$ can be modeled in numerous ways. Many of the message passing GNN architectures in the literature, such as GCN~\cite{Kipf17iclr-gcn}, GraphSAGE~\cite{Hamilton17nips-GraphSage}, GIN~\cite{Xu19iclr-gin}, GAT~\cite{Velickovic18iclr-GAT, Lee19tkdd-gat-survey}, can be used to perform message passing on $\JTH{\Graph}$. The message passing in~\eqref{eq:agg}, using edge weights, can also be made to distinguish between edges connecting to roots and children in the \Htree.
	After $\NumIterations$ iterations of message passing,
	we extract the label $y_{v}$ for node $v \in \Graph$ by combining the
	representation vectors of leaf nodes $l$ of $\JTH{\Graph}$, which correspond to node $v$ in $\Graph$, \ie $v = \kappa(l)$:
	\begin{equation}\label{eq:comb}
		y_{v} = \COMB\left(\{\vh^{\NumIterations}_{l}~|~l~\text{leaf node in $\JTH{\Graph}$~s.t.}~\kappa(l) = v\}\right),
	\end{equation}
	for every $v \in \Nodes$, where $\vh^{\NumIterations}_{l}$ denotes the representation vector generated at leaf node $l$ in $\JTH{\Graph}$ after $T$ iterations. $\COMB$ can be modeled by using any of the standard neural network models. In our experiments, we model $\COMB$ with a mean pooling function followed by a softmax.
	\omit{This neural tree architecture is shown in Figure~\ref{fig:jnt-node-classif-architecture-illlustrate}.}
	%

	
	
	\begin{remark}[Mutatis mutandis]
		The neural tree architecture is partly inspired by the junction tree algorithm~\cite{Jordan02book}. The junction tree message passing algorithm can be described in three steps. First, the clique potentials are computed for all nodes in the junction tree $(\Tree, \calB)$ of $\Graph$. This is followed by message passing between nodes on $\Tree$, which updates the clique potentials, until convergence. Third, the marginals are computed for each node from the clique potentials.
		The proposed neural tree can emulate these three steps by message passing from leaf nodes to the root nodes in \Htree, message passing between the root nodes, and message passing back from the root nodes to the leaf nodes, respectively.
		%
		%
		\cite{Xu20iclr-GNN-algoAlign} suggests that such algorithmic alignment of the neural architecture leads to \omit{lower sample complexity and} better generalizability. We leave the question of generalization power to future work. 
	\end{remark}

	

	\begin{remark}[Scalability and trade-offs]\label{rmk:scalability}
		The proposed architecture requires constructing the \Htree for the graph $\Graph$, which involves computing a \treeDecomposition of \Graph. The time and space complexity of computing a \treeDecomposition of a graph \Graph scales exponentially in the treewidth of \Graph. In many semi-supervised node classification problems, the treewidth of the input graph is too large to compute a \treeDecomposition (eg. graphs arising in citation networks~\cite{Yang16icml-revisit}). \omit{Citation networks such as Pubmed, Citeseer, Cora are some examples.}
		In such cases, to regain computational tractability, one can sub-sample the input graph (\ie remove some edges in \Graph)
		to get a graph $\Graph_{s}$ with smaller treewidth, and then apply the
		neural tree architecture to this sub-sampled graph $\Graph_s$.
		\cite{Yoo20wsdm-graphSubsampling} proposes one such graph sub-sampling algorithm, which for any given graph \Graph and a treewidth bound $k$, efficiently generates the sub-sampled graph $\Graph_s$ and its \treeDecomposition. The complexity of this algorithm is $\calO(|\Edges(\Graph)|( k^{2} + |\Nodes(\Graph)|))$.
		This addition to the neural tree architecture makes it scalable to large graphs (see Section~\ref{sec:experiments-citations}).
		%
		\omit{In Section~\ref{sec:experiments-citations}, we demonstrate that we can apply the neural tree architecture, with bounded treewidth sampling, to a citation network graph that has close to $20,000$ nodes, and we show that the choice of treewidth in the
			sub-sampled graph has negligible impact on performance.} 
		%
	\end{remark}

	\omit{
		\begin{remark}[Scalability on 3D Dynamic Scene Graphs]
			\cite{Armeni19iccv-3DsceneGraphs, Rosinol20rss-dynamicSceneGraphs} build a hierarchical 3D scene graph for actionable perception. In it, nodes are organized in a hierarchy of - object nodes, room nodes, and building nodes. A room node is connected to all the objects in the room, represented as object nodes, and a building node is connected to all the rooms, represented as room nodes. We show that such hierarchical graphs are amenable to an efficient construction of the \Htree, thereby obviating the need for any graph sub-sampling. See \suppMatt for the details.
		\end{remark}
	}

	\section{Expressive Power of Neural Trees}
	\label{sec:approximation_results}
	
	
	We now show that \NTslong can learn any graph-compatible function provided it is smooth enough.

	%
	%
	For simplicity,
	let the input node features and the representation vectors be real numbers, \ie $\vxx_v \in \mathbb{\Root}$ and $\vh^{t}_u \in \mathbb{\Root}$ for all $v \in \Nodes(\Graph)$ and nodes $u \in \JTH{\Graph}$.
	Let us implement the aggregation function $\AGG_{t}$ in~\eqref{eq:agg} as a shallow network:
	\begin{multline}
		\vh^{t}_{u} = \AGG_t\left(\vh^{t-1}_{u}, \{ (\vh^{t-1}_{w}, \kappa_{w, u})~|~w \in \nbr{u}{\JTH{\Graph}} \}\right), \\ = \textstyle{\ReLU{\sum_{k=1}^{N_u} a_{u, t}^{k}\inner{\vw_{u, t}^{k}}{\vh^{t-1}_{\bar{\calN}(u)}} + b_{u, t}^{k} }},
		\label{eq:theory_model_Agg}
	\end{multline}
	where $\bar{\calN}(u) = \{ u\}\cup \calN_{\JTH{\Graph}}(u)$ denotes the set containing node $u$ and its neighbors in the \Htree $\JTH{\Graph}$, and $a^{k}_{u, t}, b^{k}_{u, t}, \vw_{u, t}^{k},$ and $N_u$ are parameters.\footnote{We assume a different $\AGG_t$ function for each node $u$ at iteration $t$. This choice is more general than our architecture in Section~\ref{sec:architecture-message-passing}. However, our results extend to the case where the $\AGG_{t}$ function is the same across nodes $u$ in each iteration $t$.} The representation vectors $\vh^{t}_u$ are initialized as discussed in Section~\ref{sec:architecture-message-passing}.
	We fix a node $v_{0}$ in graph $\Graph$ and extract our output from $v_0$:
	\begin{equation}\label{eq:theory_model_comb}
		\textstyle{y_{v_0} = \COMB\left(\{\vh^{\NumIterations}_{l}~|~l~\text{leaf node in $\JTH{\Graph}$~s.t.}~\kappa(l) = v_0\}\right),}
	\end{equation}
	where $T$ is the number of iterations.
	We also assume the $\COMB$ function to be a shallow network. 
	%
	%
	Consider the space of  
	functions {$g$} that map the input node features $\MX$ to the output $y_{v_0}$ (in~\eqref{eq:theory_model_comb}):
	\begin{equation}
		\nonumber
		\calF(\Graph, N) = \left\{ g :\MX \rightarrow g(\MX) = y_{v_0}~\!\Bigg|\!\begin{array}{c}
			\text{For some}~T > 0~\text{s.t.} \\
			y_{v_0}~\text{given by~\eqref{eq:theory_model_Agg}-\eqref{eq:theory_model_comb}}
		\end{array}\!\!\!\right\},
	\end{equation}
	where $N$ denotes the total number of parameters used in the \NTlong architecture.
	
	We now show that any graph-compatible function -- that is smooth enough -- can be approximated by a function in $\calF(\Graph, N)$ to an arbitrary precision.
	\begin{theorem}
		\label{thm:approx}
		Let $f:[0,1]^{n} \rightarrow [0, 1]$ be a function compatible with a graph $\Graph$ with $n$ nodes. Let each   clique function $\theta_{c}$ in $f$ (see Definition~\ref{def:compatible_fun}) be $1$-Lipschitz and be bounded to $[0, 1]$.
		Then, for any $\epsilon > 0$, there exists a $g \!\in\! \calF(\Graph, N)$ such that $|| f \!-\! g||_{\infty} \!<\! \epsilon$, while the number of parameters $N$ is bounded by
		\begin{equation}
			\label{eq:param_approx}
			N = \textstyle \calO\left( \sum_{u \in \Nodes(\JTH{\Graph})} (d_u-1) \left( \frac{\epsilon}{d_u-1} \right)^{-(d_u-1)}\right),
		\end{equation}
		where $d_u$ denotes the degree of node $u$ in $\JTH{\Graph}$, and
		the summation is over all the non-leaf nodes in $\JTH{\Graph}$.
	\end{theorem}
	\begin{IEEEproof}
		See Appendix~\ref{sec:app-approximation}.
	\end{IEEEproof}
	
	\begin{remark}[Bounded Functions]
		Theorem~\ref{thm:approx} assumes the domain and range of the compatible function $f$ and the clique functions $\theta_c$ to be bounded between $[0, 1]$. We remark here that the result, and the proof, can be extended to any bounded $f$ and $\theta_c$, over bounded domains.
	\end{remark}
	
	We next develop the bound in Theorem~\ref{thm:approx}
	to expose the dependence of the number of parameters $N$ on the treewidth of the tree decomposition of the graph.
	
	\begin{corollary}
		\label{cor:approx}
		The number of parameters $N$ in Theorem~\ref{thm:approx} is upper-bounded by
		\begin{equation}
			\nonumber
			N = \calO\left( n \times (\treewidth{\JTH{\Graph}} + 1)^{2\treewidth{\JTH{\Graph}} + 3}\times \epsilon^{- (\treewidth{\JTH{\Graph}} + 1)}\right),
		\end{equation}
		where $\treewidth{\JTH{\Graph}}$ denotes the treewidth of the tree-decomposition of $\Graph$, formed by the root nodes of $\JTH{\Graph}$.
		
	\end{corollary}
	\begin{IEEEproof}
		See Appendix~\ref{sec:app-bound}.
	\end{IEEEproof}

	\begin{remark}[Efficient approximations]
		Corollary~\ref{cor:approx}~shows that the number of parameters needed to obtain an $\epsilon$-approximation with \NTslong increases exponentially in only the treewidth of the tree-decomposition, and is linear in the number of nodes $n$ in the graph. Thus, for graphs with bounded treewidth, \NTslong are able to approximate any graph-compatible function efficiently.
	\end{remark}
	
	\omit{\begin{remark}[Comparing with shallow and deep neural networks]
			If we were to take a single-layer neural network, as in~\eqref{eq:theory_model_Agg}, that aggregates all the input features $\MX$ to approximate $f$ without any regard to its compatibility structure, we would need $N = \mathcal{O}\left( \epsilon^{-n}\right)$ number of parameters to achieve the same $\epsilon$ approximation~\cite{Poggio2017j-IJAC-WhyWhenDNN}. In this case, we see $N$ growing exponentially in the graph size $n$.
			It can be argued that the same advantage is shared by a deep neural network of depth $\treewidth{\JTH{\Graph}}$~\cite{Poggio2017j-IJAC-WhyWhenDNN}. However, a general multi-layer perceptron would have to train many more parameters (a constant factor more) to first zone in on the exact message passing structure for computing the compatible function $f$. However, in our case, this is given by the construction of the \Htree.
	\end{remark}}
	
	\begin{remark}[Data efficiency]
		\label{rem:data-efficiency}
		The value of $N$ also affects the data required to train the model: the larger the $N$, the more samples
		are required for training. 
		In particular, Corollary~\ref{cor:approx} provides the reassuring result that
		if the training dataset contains graphs of small treewidth,
		then the amount of data required for training scales only linearly in the number of nodes $n$.
	\end{remark}

	\section{Experiments}
	\label{sec:experiments}
	
	This section shows that the neural tree architecture outperforms standard Graph Neural Network architectures on 3D Scene Graph node classification (Section~\ref{sec:experiments-3DSceneGraph}).
	%
	We then demonstrate the practicability of the neural tree architecture on much larger citation network datasets and show how its implementation ---along with the bounded treewidth subgraph sampling algorithm in~\cite{Yoo20wsdm-graphSubsampling}--- leads to improved performance, even for small treewidht bounds (Section~\ref{sec:experiments-citations}).
	
	\subsection{Node Classification in 3D Scene Graphs}
	\label{sec:experiments-3DSceneGraph}
	We use the neural tree architecture for node classification on 3D scene graphs and show it outperforms the standard GNNs.
	
	\textbf{Dataset.} We run semi-supervised node classification 
	\begin{wrapfigure}[15]{r}{0.28\textwidth}
		\vspace{-0.1cm}
		\centering
		\includegraphics[trim=8 30 20 20,clip, width=0.95\linewidth]{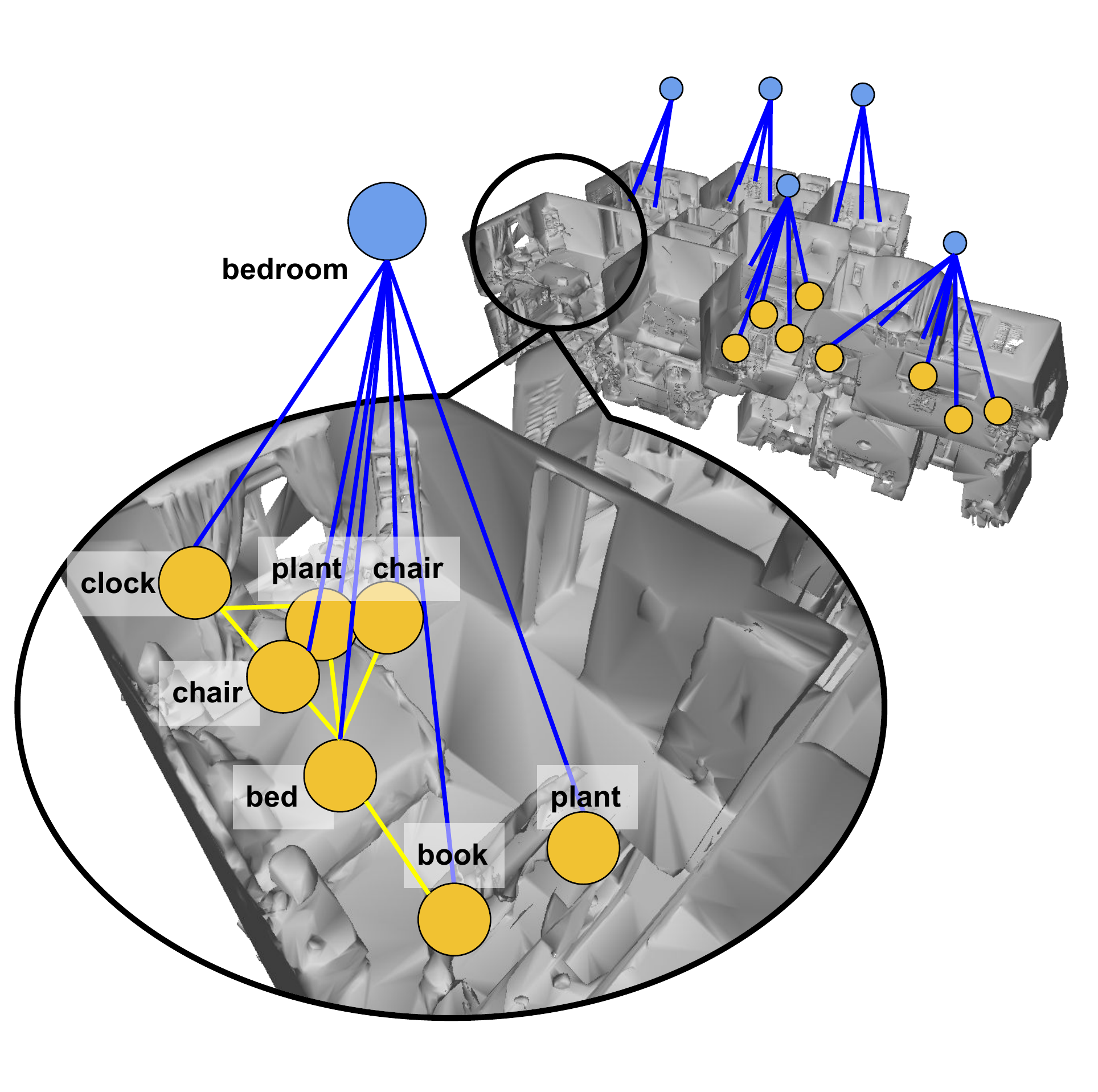}
		\caption
		{A room-object graph in a 3D scene graph.%
			\label{fig:3DSceneGraph}
		}
	\end{wrapfigure}
	experiments on Stanford's 3D scene graph dataset ~\cite{Armeni19iccv-3DsceneGraphs}. The dataset includes 35 3D scene graphs with verified semantic labels, each containing building, room, and object nodes in a residential unit.
	Since there is only a single class of building nodes (residential),
	we remove the building node and obtain 482 room-object graphs where each graph contains a room and at least one object in that room as shown in Fig.~\ref{fig:3DSceneGraph}. The resulting dataset has 482 room nodes with 15 semantic labels, and 2338 objects with 35 labels. Each object node is connected to the room node it belongs to. In addition we add 4920 edges to connect adjacent objects in the same room.
	We use the centroid and bounding box dimensions as features for each node.

	\myParagraph{Approaches and Setup}
	We implement the neural tree architecture with
	four different aggregation functions $\text{AGG}_t$ specified in: GCN~\cite{Kipf17iclr-gcn}, GraphSAGE~\cite{Hamilton17nips-GraphSage}, GAT~\cite{Velickovic18iclr-GAT}, GIN~\cite{Xu19iclr-gin}. We randomly select 10\% of the nodes for validation and 20\% for testing. 
	The hyper-parameters of the two approaches are separately tuned based on the best validation accuracy, while using all 70\% of the remaining nodes for training; see
	Appendix~\ref{sec:app-experiments} for details.
	\omit{We use the ReLU activation function and also implement dropout at each iteration.}
	\omit{The $\text{READ}$ function for the standard GNN (see~\eqref{eq:GNN_READ}) is implemented as a single linear layer followed by a softmax.
		On the other hand, the $\text{COMB}$ function (see~\eqref{eq:comb}) for neural trees is implemented as a mean pooling operation, followed by a single linear layer and a softmax. We use different READ (resp. COMB) functions for the room nodes and the object nodes.
		We train the architectures using the standard cross entropy loss function. The experiments are implemented using the PyTorch Geometric library.}
	%
	%
	%

	%
	\omit{We compare the proposed neural tree architecture with the standard GNN architectures, which perform message passing on the input graph.} 
	\begin{table}[t]
		\centering
		\caption{Test Accuracy}
		\label{tab:sg_results}
		\scalebox{1.00}{
			\begin{tabular}{lll}
				\toprule
				Model           & Input graph               & {Neural Tree\xspace} \\
				\midrule
				GCN  	        & $40.88 \pm 2.28$ \% & ${\bf 50.63} \pm 2.25$ \%	\\
				GraphSAGE       & $59.54 \pm 1.35$ \% & ${\bf 63.57} \pm 1.54$ \% \\					
				GAT             & $46.56 \pm 2.21$ \% & ${\bf 62.16} \pm 2.03$ \% \\
				GIN             & $49.25 \pm 1.15$ \% & ${\bf 63.53} \pm 1.38$ \% \\
				\bottomrule
		\end{tabular}}
		\vspace{-0.5cm}
	\end{table}
	\textbf{Results.} Table~\ref{tab:sg_results}  compares the test accuracies (averaged over 100 runs) for the standard GNN architectures and the corresponding neural tree architecture, while using the same type of aggregation function.
	\omit{The reported test accuracies are averaged over 100 runs.}
	We see that the neural tree architecture always yields a better prediction model than the standard GNN, for a given aggregation function.

	To further analyze the proposed architecture, we carry out a series of experiments to see how the test
	accuracy varies as a function of the amount of training data and the number of message passing iterations $T$.
	%
	%
	For simplicity, we only show the neural tree that uses the GCN aggregation function in comparison with the standard GCN.

\begin{figure}[h]
		\centering
		\vspace{-0.6cm}
		\begin{minipage}[b]{0.49\linewidth}
			\includegraphics[trim=45 135 25 150,clip,width=\linewidth]{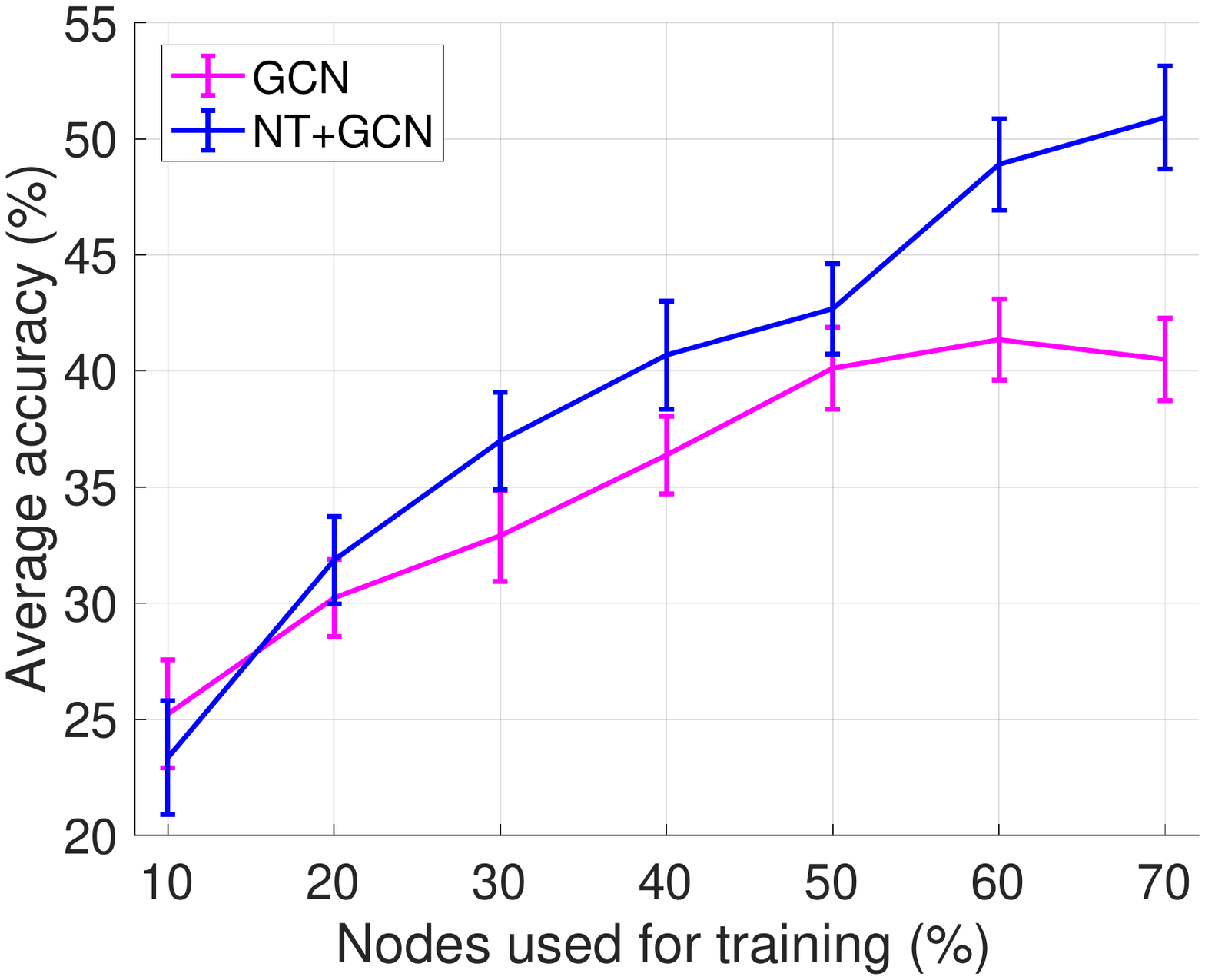}
			\caption{Accuracy vs. training data (\% of labeled nodes).}
			\label{fig:ratio}
		\end{minipage}
		\hfill
		\begin{minipage}[b]{0.49\linewidth}
			\includegraphics[trim=45 140 25 150,clip,width=\linewidth]{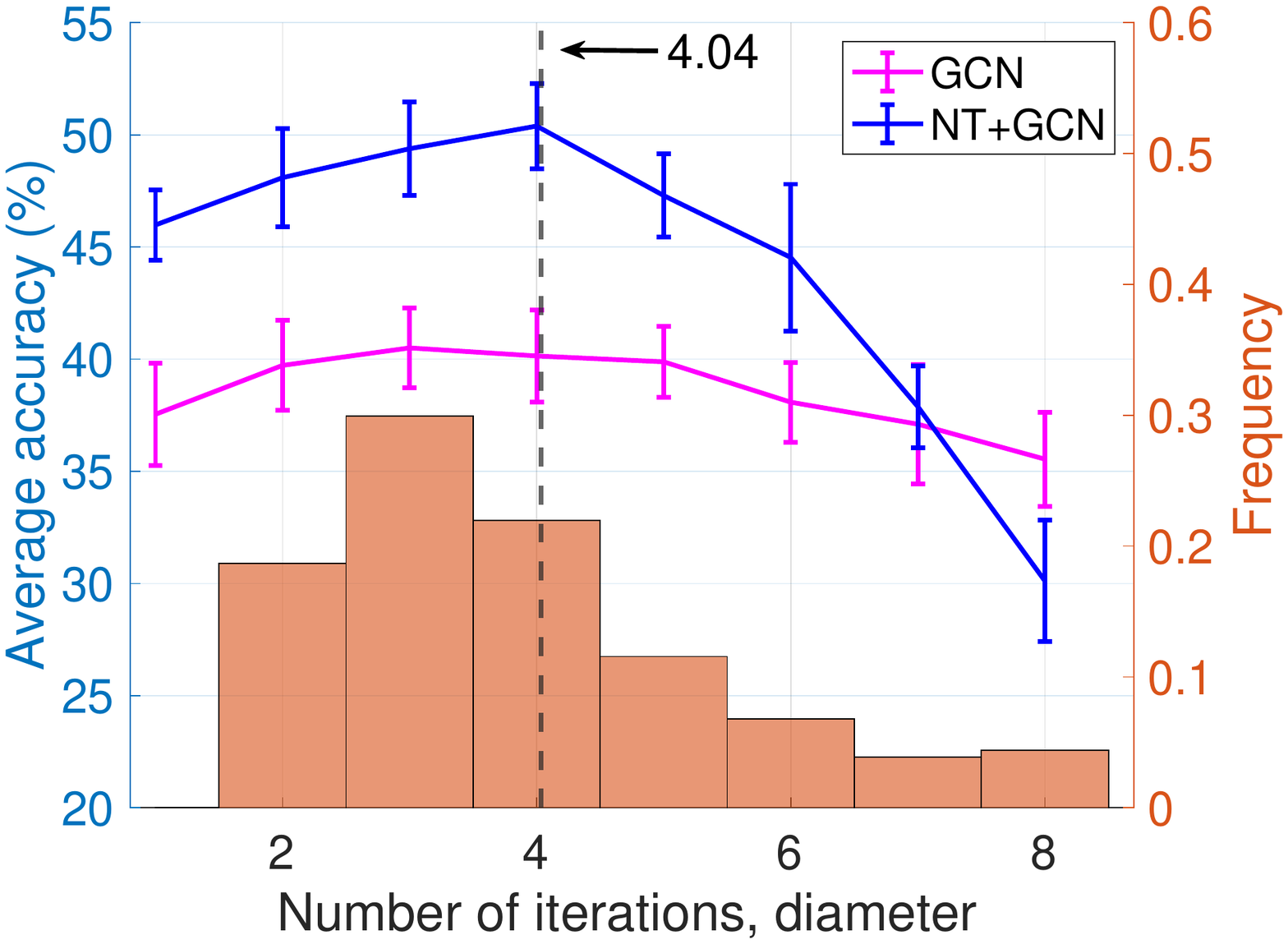}
			\caption{Accuracy vs. number of iterations and weighted diameter distribution.}
			\label{fig:iterations}
		\end{minipage}
	\end{figure}

	Figures~\ref{fig:ratio} and~\ref{fig:iterations} plot the test accuracy (averaged over 10 runs) as a function of the training data used and the number of iterations $T$.
	The test accuracy  -- for both the neural tree (NT$+$GCN) and GCN -- increases with increasing training data, however, the increase is sharper for the neural tree architecture, eventually outperforming GCN. 
	
	This shows the higher expressive power of the proposed neural tree architecture.
	\begin{figure*}
		\centering
		\begin{tabular}{ccc}
			\includegraphics[trim=45 155 45 155,clip,width=0.28\linewidth]{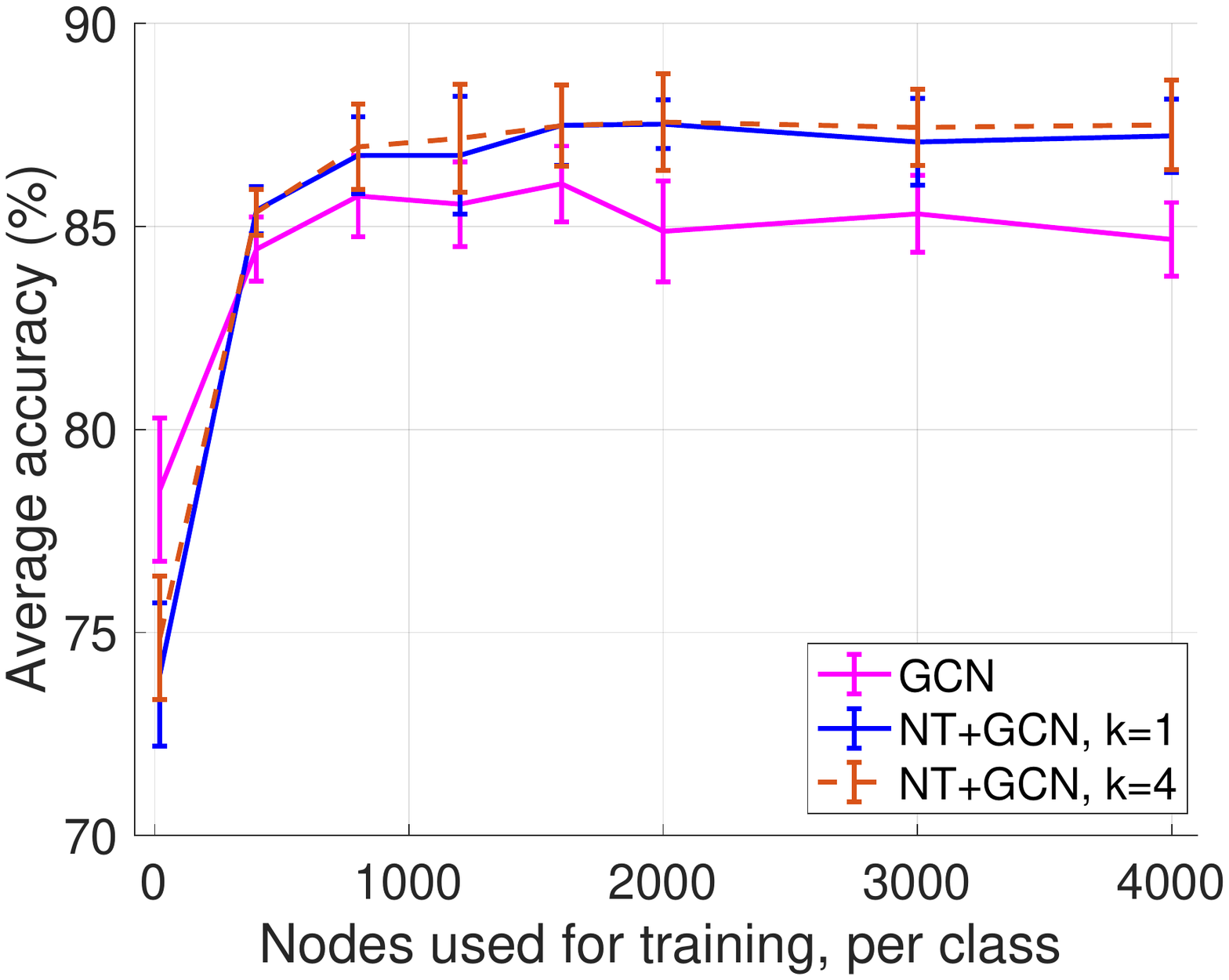} &   \includegraphics[trim=45 155 45 155,clip,width=0.28\linewidth]{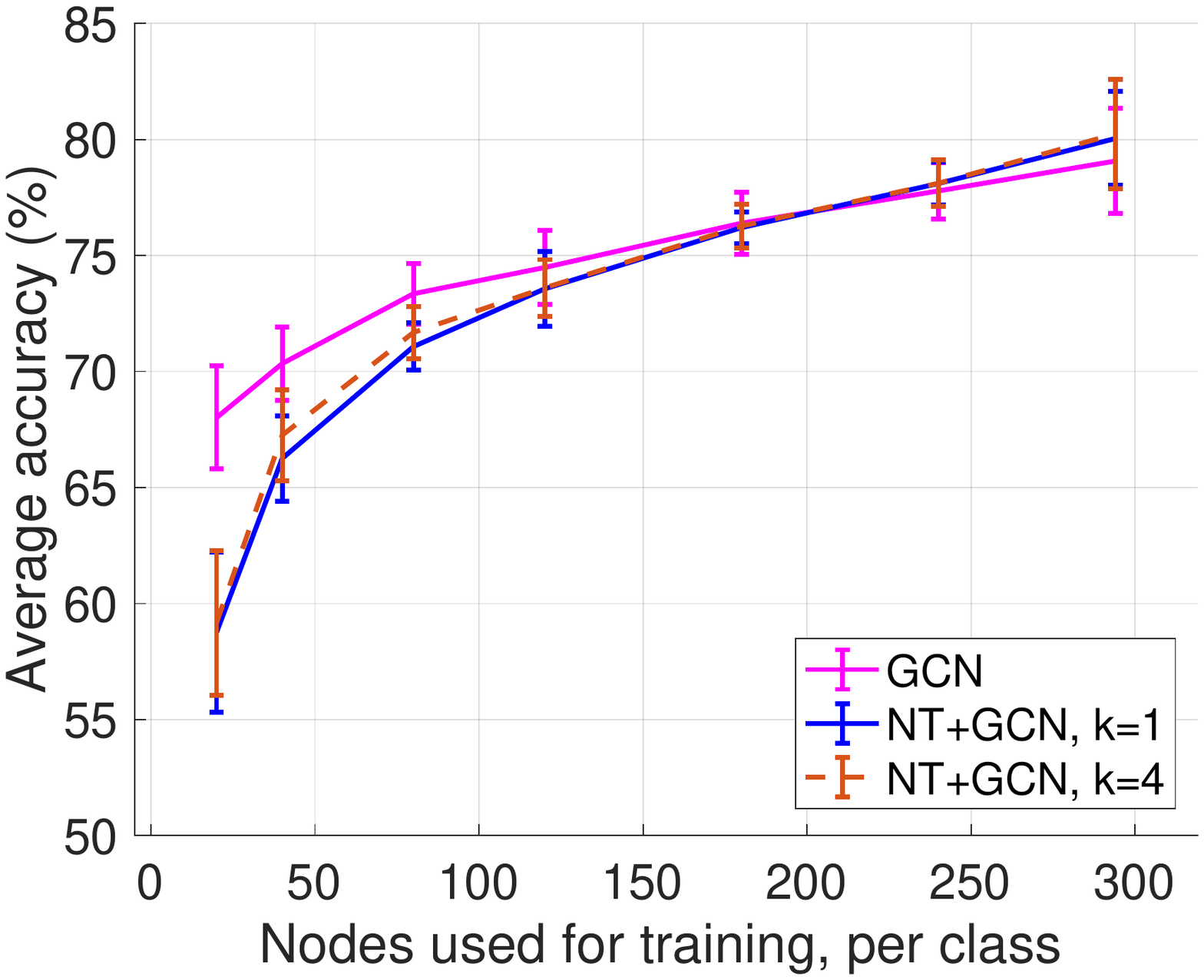} &   \includegraphics[trim=45 155 45 155,clip,width=0.28\linewidth]{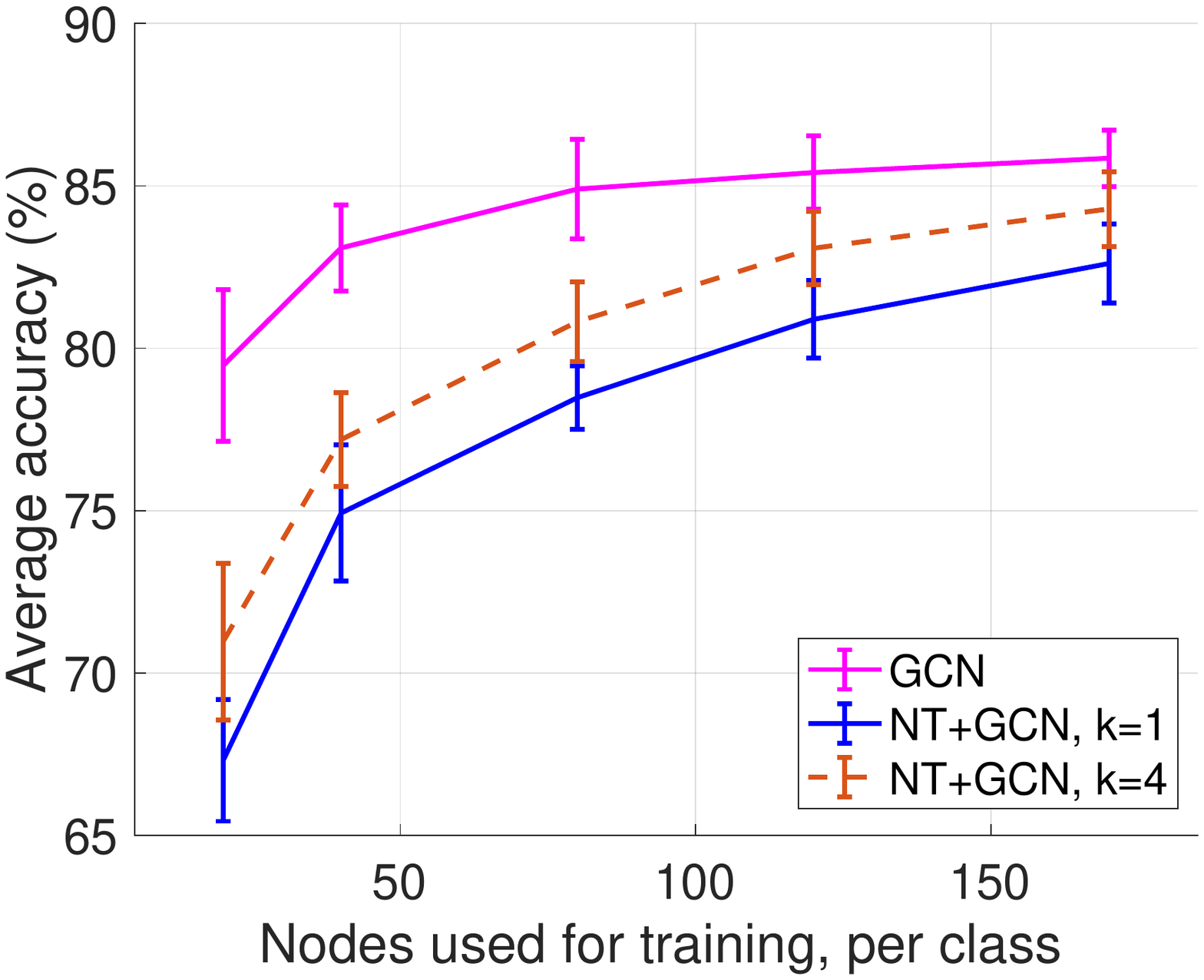} \\
			(a) NT+GCN on PubMed  & (b) NT+GCN on CiteSeer & (c) NT+GCN on Cora \\[6pt]
			\includegraphics[trim=45 155 45 155,clip,width=0.28\linewidth]{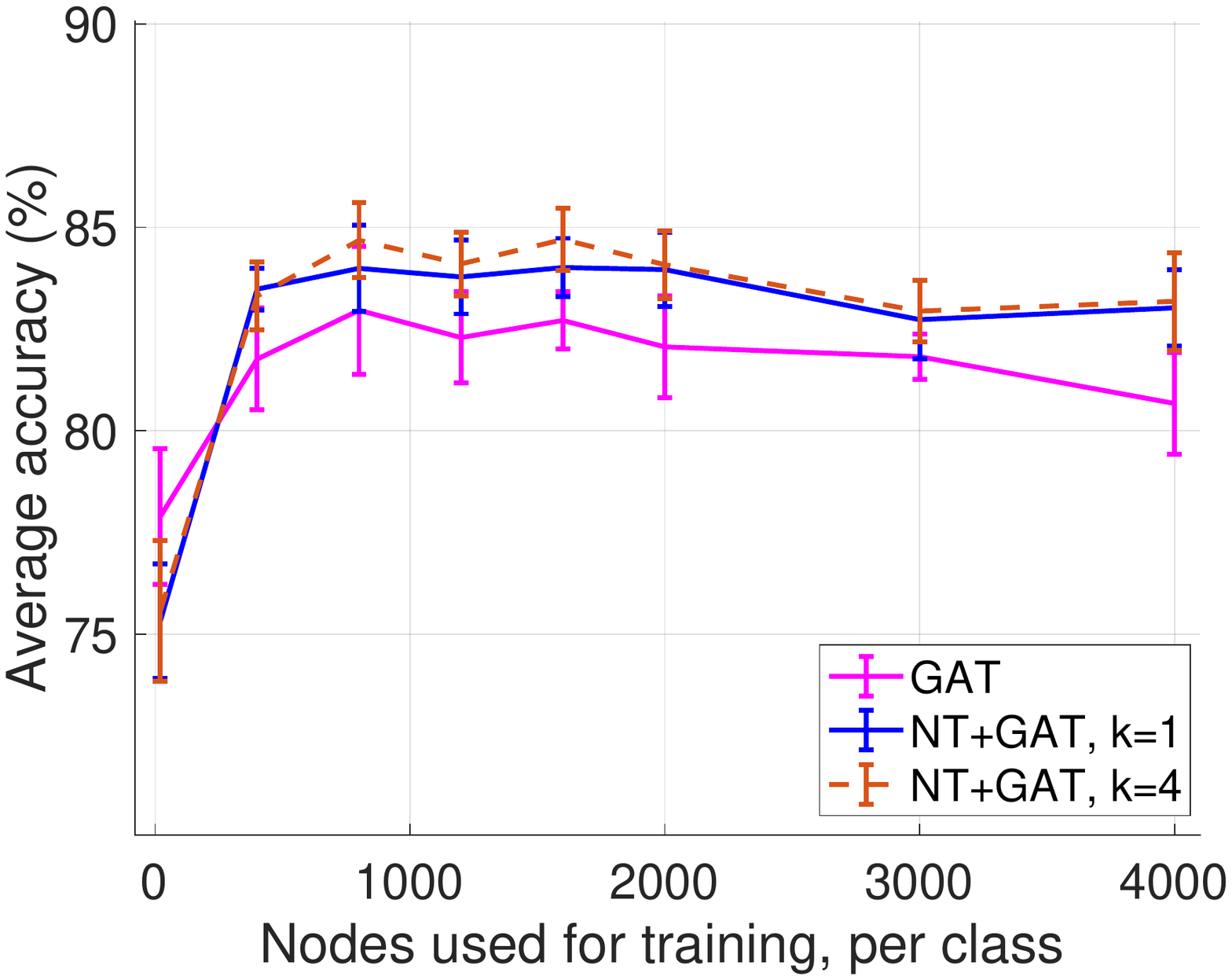} &   \includegraphics[trim=45 155 45 155,clip,width=0.28\linewidth]{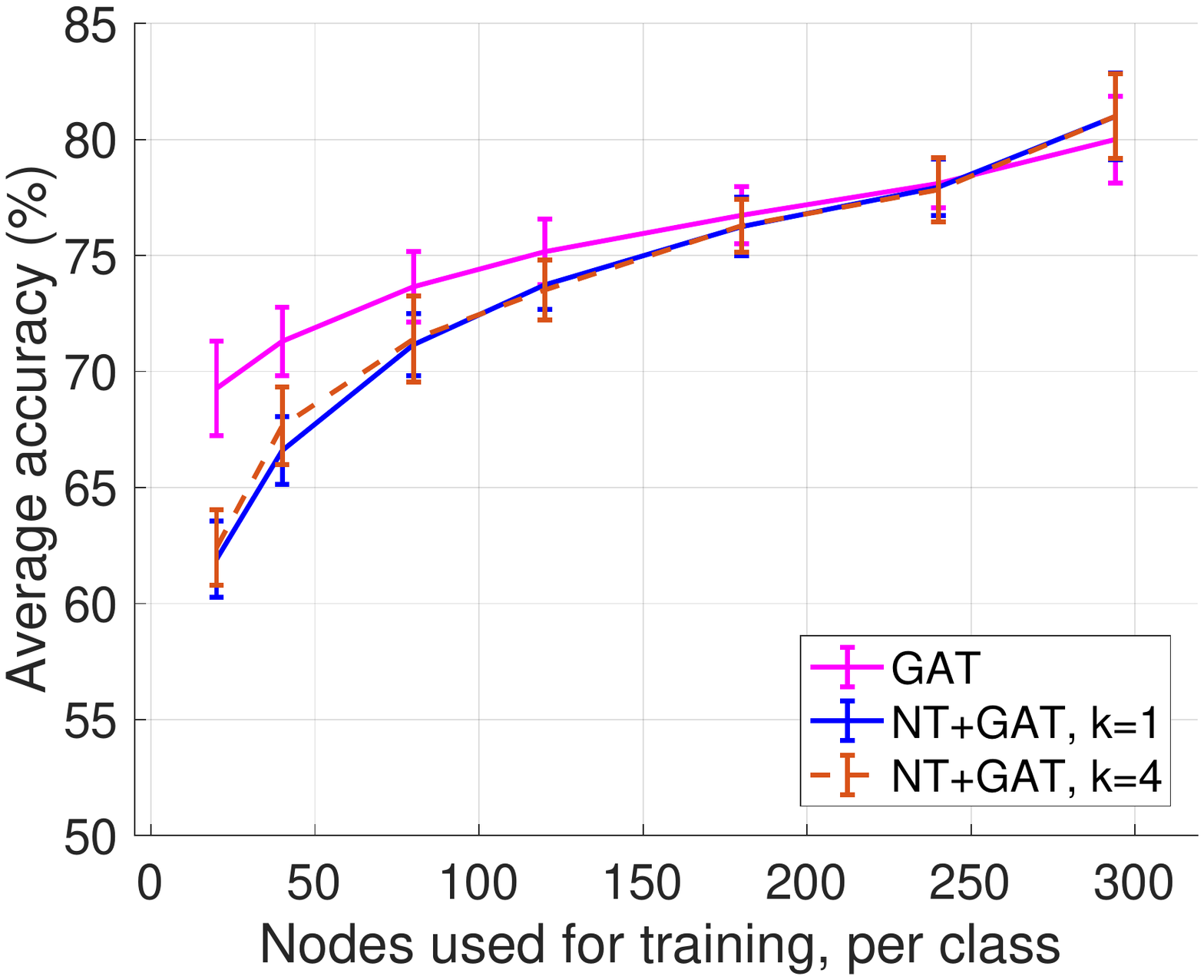} &   \includegraphics[trim=45 155 45 155,clip,width=0.28\linewidth]{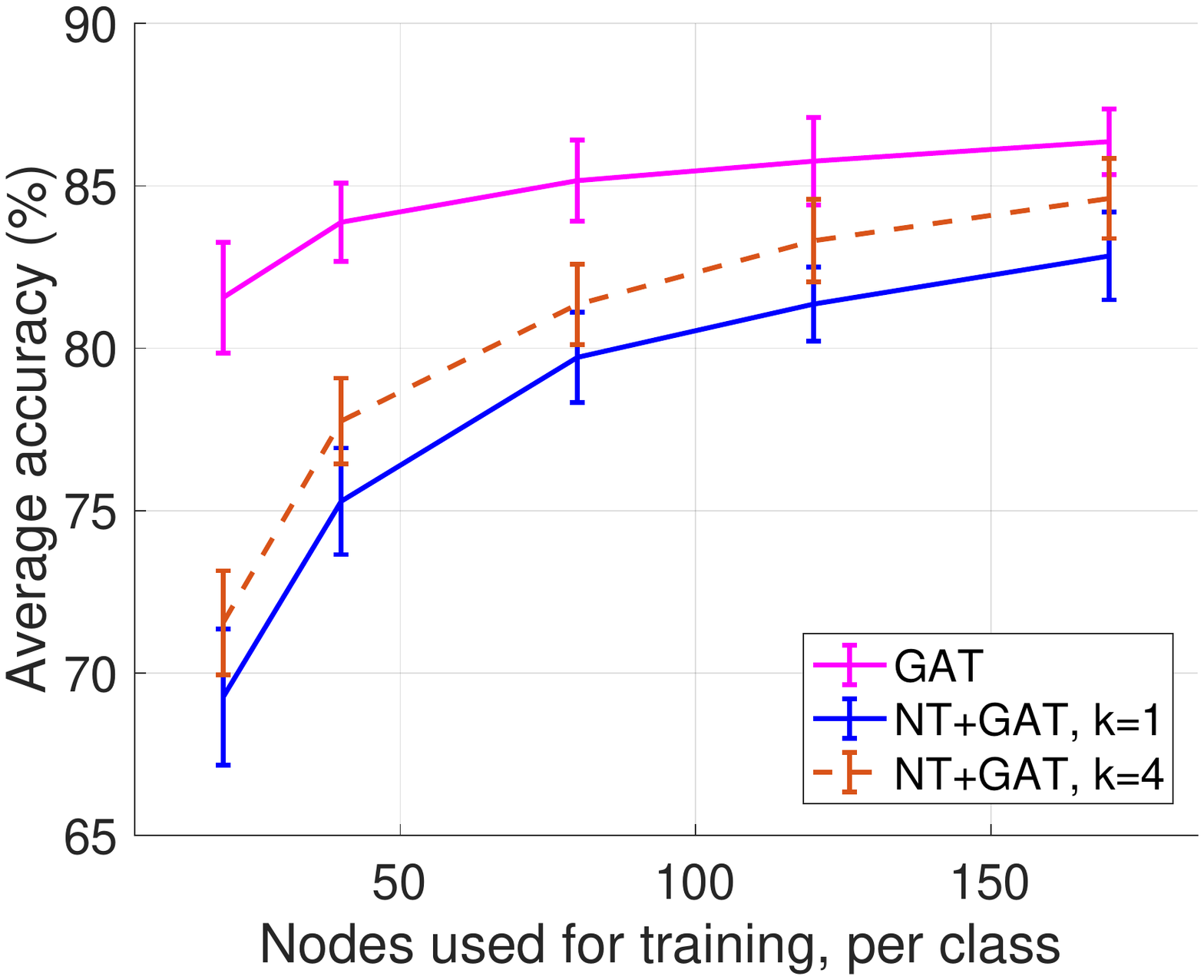}\\
			(d) NT+GAT on PubMed & (e) NT+GAT on CiteSeer & (f) NT+GAT on Cora \\[6pt]
		\end{tabular}
		\caption{Accuracy vs training nodes (per label).}
		\label{fig:accuracy-vs-data}
		\vspace{-0.2cm}
	\end{figure*}
	As with the number of iterations $T$, we see an optimal $T$ at which the test accuracy is maximized. This optimal $T$ is empirically close to the average diameter of the constructed \Htrees, of all (room-object) scene graphs in the dataset.
	This is intuitive, as for the messages to propagate across the entire \Htree, $T$ would have to equal the diameter of the \Htree. See Appendix~\ref{sec:app-experiments} 
	for more details, where we also report the compute, train, and test time requirements for neural trees.
	%

	\subsection{Node Classification in Citation Networks}
	\label{sec:experiments-citations}
	We now demonstrate the applicability of the neural tree architecture to large networks with high treewidth by using it in conjunction with the bounded treewidth subgraph sampling proposed in~\cite{Yoo20wsdm-graphSubsampling}.
	We use the popular citation network datasets~\cite{Yang16icml-revisit}, where nodes are documents and undirected edges are citations. Each node has a class label representing the subject of the document. These graphs have high treewidth, and therefore, are first sampled using the bounded treewidth subgraph sampling algorithm in~\cite{Yoo20wsdm-graphSubsampling}. The neural tree is constructed on the sampled graph.

	\textbf{Datasets} We use three popular citation network datasets ---PubMed, CiteSeer, and Cora~\cite{Yang16icml-revisit}--- where nodes are documents and undirected edges are citations. Each node has a class label representing the subject of the document. Table~\ref{dataset-statistics-table} outlines statistics about the dataset.
	\begin{table}[b]
		\vspace{-0.1cm}
		\centering
		\caption{Citation network dataset statistics~\cite{Yang16icml-revisit}.}
		\label{dataset-statistics-table}
		\vspace{-0.4cm}
		\begin{center}
			\begin{small}
				\begin{tabular}{lccccr}
					\toprule
					& PubMed   & CiteSeer 	& Cora \\
					\midrule
					Nodes 			& 19,717   & 3,327 	    & 2,708 \\
					Edges 			& 44,338   & 4,732 	    & 5,429 \\
					Classes 	    &	3      & 6 		    & 7 	\\
					\bottomrule
				\end{tabular}
			\end{small}
		\end{center}
		\vspace{-0.2cm}
	\end{table}
	The input citation network graphs have high treewidth, and therefore, are first sampled (see Remark~\ref{rmk:scalability} in Section~\ref{sec:architecture}) using the bounded treewidth subgraph sampling algorithm in~\cite{Yoo20wsdm-graphSubsampling}, with a treewidth bound of $k$. The neural tree is then constructed from the sampled subgraph.

	\myParagraph{Approaches and Setup}
	We implement the neural tree architecture with the aggregation function $\text{AGG}_t$ specified in: GCN~\cite{Kipf17iclr-gcn} and GAT~\cite{Velickovic18iclr-GAT}. The READ function (see~\eqref{eq:GNN_READ}) is implemented as a softmax, same as in~\cite{Kipf17iclr-gcn, Velickovic18iclr-GAT}, and the COMB function (see~\eqref{eq:comb}) is implemented as a mean pooling operation, followed by a softmax. See Appendix~\ref{sec:app-experiments-cite} for more details.

	\textbf{Results.}
	Figure~\ref{fig:accuracy-vs-data} plots test accuracy as a function of training data for all the three datasets. The test accuracy, for both standard GNNs and neural trees, increase with increasing number of training nodes. However, the increase tends to be much sharper for neural trees. 
	Also note that, on the PubMed dataset, the test accuracy for the neural trees settles, after the sharp increase, to a value that is above the corresponding GNN architecture ((a) and (d) in Fig.~\ref{fig:accuracy-vs-data}). However, on the CiteSeer and Cora dataset, the test accuracy never really crosses the standard GNN architecture.
	This is because the number of available training nodes (per label) is much less in the CiteSeer and Cora dataset, than it is in the PubMed dataset. 

	This indicates that the performance of neural trees is directly proportional to the amount of available training data. While the standard GNNs can be expected to perform well when there is less available training data, the neural trees will most likely perform better in the high training data regime.
	We attribute this to the higher expressive power of the neural tree architecture. \emph{The neural tree architecture is able to seep in more data to yield higher prediction accuracy.}
	
	\begin{figure}[t]
		\vspace{-0.5cm}
		\begin{tabular}{cc}
			\includegraphics[trim=45 170 45 170,clip,width=0.49\linewidth]{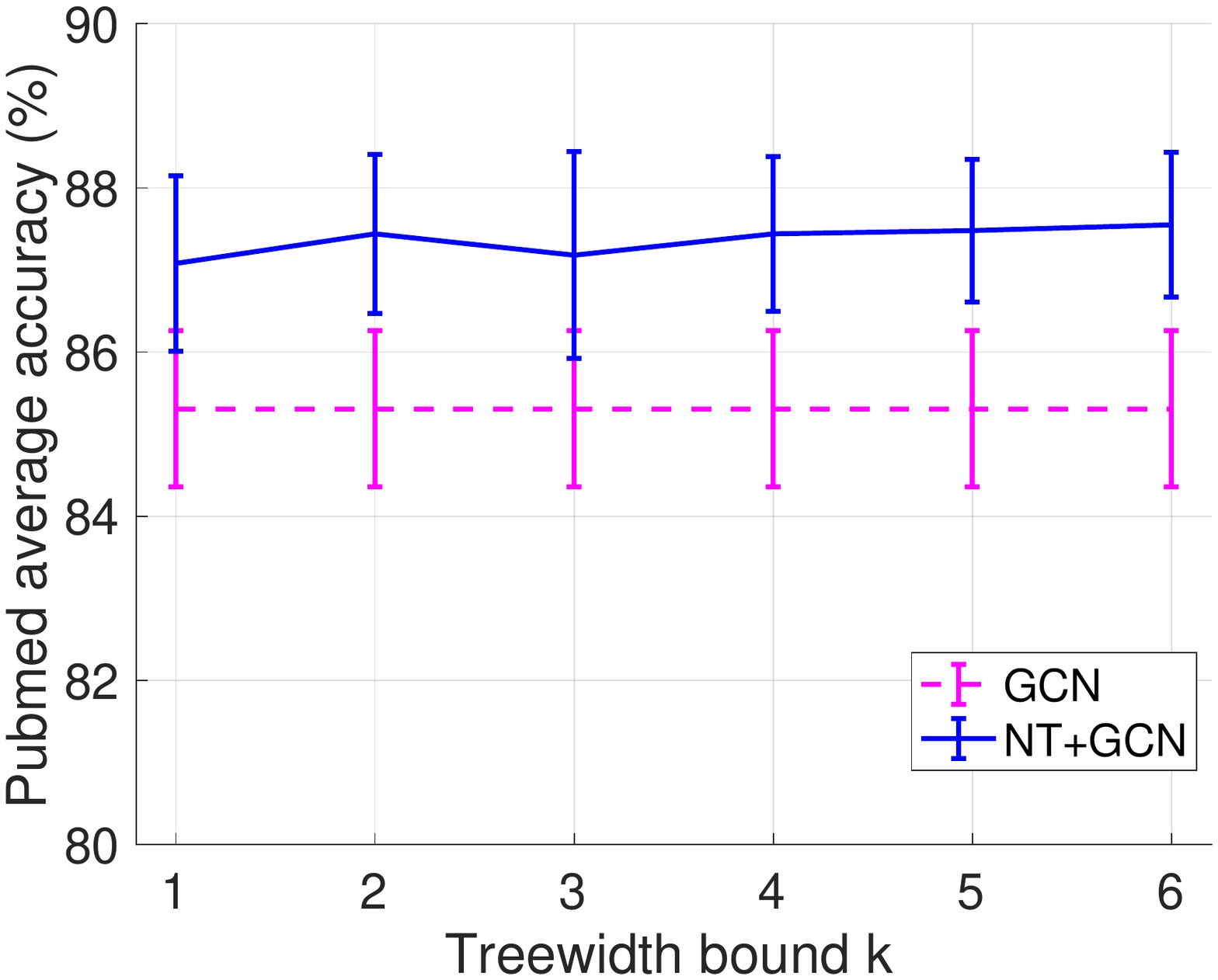} &   \includegraphics[trim=45 170 45 170,clip,width=0.49\linewidth]{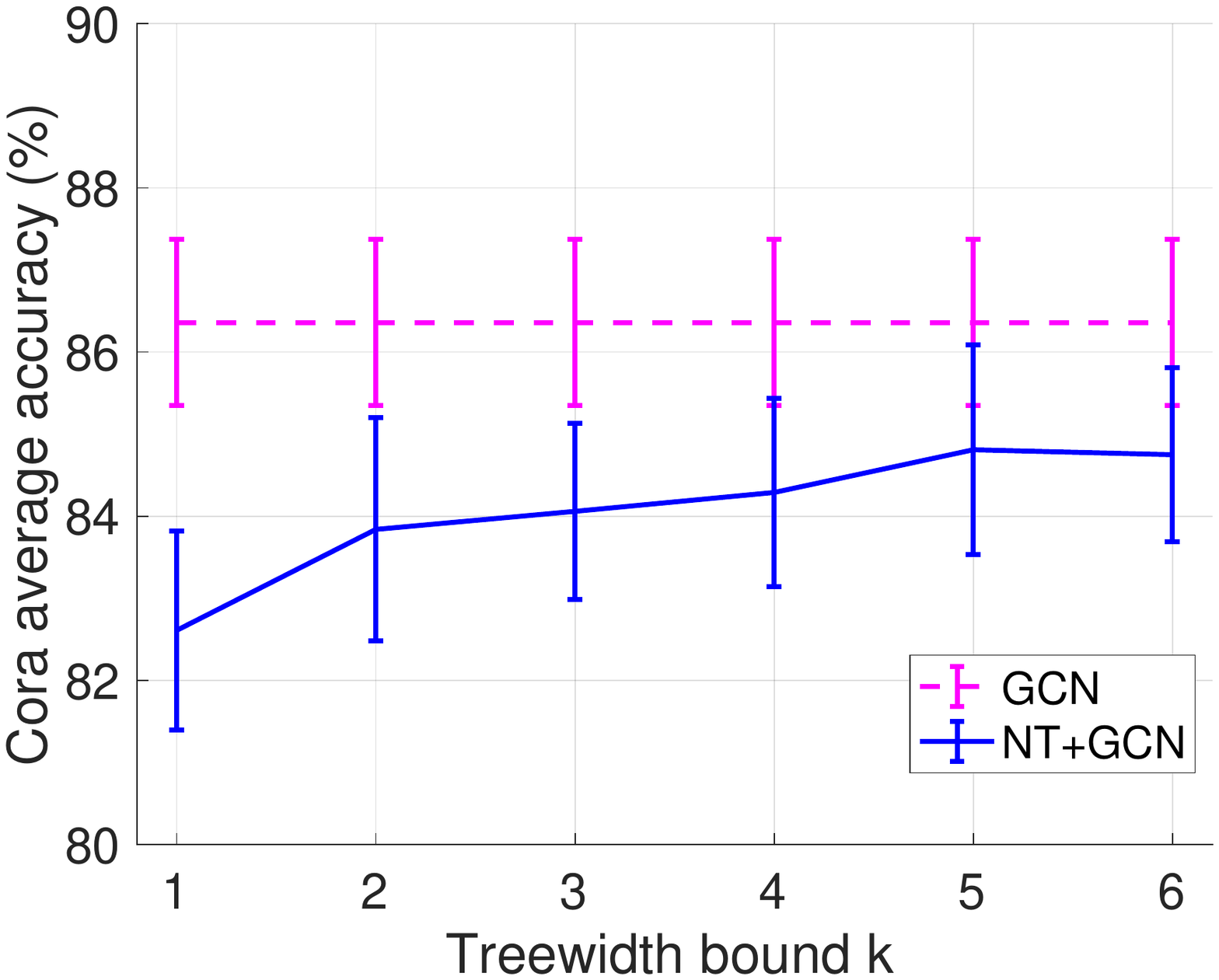} \\
			(a) PubMed & (b) Cora \\[6pt]
			\vspace{-0.5cm}
		\end{tabular}
		\caption{Average accuracy as a function of treewidth bound $k$ for NT+GCN.}
		\label{fig:tw_eb}
		\vspace{-0.3cm}
	\end{figure}
	
	The most noticeable element in Figure~~\ref{fig:accuracy-vs-data} is the variation (or lack of it) in prediction accuracy in the treewidth bound $k$. Recall that the input graph is first sampled using the bounded treewidth subgraph sampling algorithm from~\cite{Yoo20wsdm-graphSubsampling} (see Remark~\ref{rmk:scalability} in Section~\ref{sec:architecture}). On the PubMed and CiteSeer dataset, we observe that the treewidth bound $k$ used for subgraph sampling does not have much of an effect on the prediction accuracy. However, on Cora dataset, the performance can be improved by increasing the treewidth bound $k$. To further investigate this, we plot the average test accuracy as a function of treewidth bound $k$ in Figure~\ref{fig:tw_eb} (for NT+GCN, on PubMed and Cora). 
	We observe that while the prediction accuracy remains nearly the same on PubMed, there is a noticeable increase on Cora.
	
	This indicates that in some datasets (\eg PubMed, CiteSeer) it is possible to retain the best possible performance, even after sampling the input graph with a very low treewidth bound; say $k=1$. This is very significant as it means that even if we disregard many of the existing edges in the network dataset, the performance does not degrade much. In the case of other datasets (\eg Cora), choosing a low treewidth bound $k$ serves as a good approximate solution. Note that the test accuracy gap between $k=6$ and $k=1$ is only about $2$ percentage points, in Cora (see Figure~\ref{fig:tw_eb}).
	
	These results show that \emph{neural tree is a scalable architecture and can be applied to large networks with high treewidth}. The choice of the treewidth bound $k$ will have to be tailored to the dataset in question. However, in order to achieve the full expressive power of neural trees, more training data is required.

	\section{Conclusion}
	\label{sec:conclusion}
	
	
	We propose a novel graph neural network architecture -- the \emph{neural tree}.
	The neural tree performs
	message passing, not on the input graph, but over a constructed \Htree, which
	provides a tree-structured description of the original graph and its subgraphs.
	We show that the neural tree architecture can approximate any graph-compatible function,
	and that the number of parameters required to obtain a desired approximation grows linearly
	with the number of nodes and exponentially in the treewidth of the input graph.
	This renders the proposed architecture more parsimonious for large graphs with small treewidth.
	
	Graph-compatible functions arise in probabilistic graphical models, hence the proposed architecture can approximate any probability distribution function defined on a graph.
	%
	%
	%
	Furthermore, we show that a graph-compatible function can be used to approximate any smooth graph-invariant/equivariant functions
	studied in the literature.
	This suggests that the goal of approximating graph-compatible functions is a worthwhile pursuit towards the
	design of novel GNN architectures.
	
	We use neural trees for node classification on 3D scene graph and citation network datasets, showing that the proposed architecture leads to
	more accurate predictions with increasing training data and is applicable even for large networks with high treewidth. 
	
	\emph{Neural Tree} is a general purpose architecture and remains to be applied to other learning tasks such as graph representation learning and classification. \omit{As a future work, we believe, it interesting to explore the connection between \Htree and the neural network architectures that seek to extract hierarchical or subgraph-level features in graph learning.}

	\section{Societal Impact}
	\label{sec:impact}

	\myParagraph{Research Community} Many problems have been sought to be solved using graph neural networks. However, the relation between complexity of the underlying problem and parameter complexity of the neural architecture used to solve it is not generally well investigated. Moreover, the expressivity of the graph neural network architecture, \ie, its ability to solve any instance of the problem is also not fully understood. 
	
	This work, we believe, is a step towards understanding these fundamental questions. In obtaining approximation guarantees for the proposed \emph{Neural Tree} architecture, we bring out an interesting tangle between approximating graph compatible functions (which can be thought of as approximating exact inference over probabilistic graphical models), graph treewidth, and the parameter complexity of the Neural Tree. The parameter complexity obtained in Theorem~\ref{cor:approx} matches the problem complexity of exact inference on probabilistic graphical models~\cite{venkat12uai}.
	
	We hope that this work will inspire other researchers to consider similar questions - for other problems and neural architectures - and investigate the relation between the problem complexity, parameter complexity, and the underlying graph properties - such as the graph treewidth.

	\myParagraph{Community at Large} The main thrust of this work is to develop a new graph neural network architecture that can approximate any graph compatible function. We show that the parameter complexity increases exponentially in the graph treewidth, and is of the same order as the complexity of exact inference on graphical models. This implies that when applying Neural Trees, graph treewidth is not only an important parameter, but the most important aspect in controlling the required memory and computation time. 
	
	This can be a limiting factor in deploying the Neural Tree architecture in cases where either the energy consumption or large graph treewidth is an issue. In the paper, however, we observe that using the Neural Tree architecture in conjunction with bounded treewidth subgraph sampling~\cite{Yoo20wsdm-graphSubsampling} provides a good approximation in such cases.

	\section{Acknowledgment}	
	This work was partially funded by the Office of Naval Research under the ONR RAIDER program (N00014-18-1-2828).

	\bibliographystyle{ieeetr}

	\appendix


\subsection{Notations}
\label{sec:app-notations}
All graphs in this paper are undirected and simple, \ie they do not contain multiple edges between  two nodes.
For a graph \Graph, we also use $\Nodes(\Graph)$ and $\Edges(\Graph)$ to denote the set of nodes and edges, respectively, and $n$ to denote the number of nodes, namely $n = |\Nodes(\Graph)|$.
We use $\Graph[\setA]$ to denote the subgraph of $\Graph$ induced by a subset of nodes $\setA \subset \Nodes(\Graph)$. $|\setA|$ denotes the size of the set $\setA$.
%
%
As described, we use $\vxx_v$ to denote the feature vector of $v$, while
the space of all node features of $v$ is denoted by $\setX_v$ and $\times_{v \in V} \setX_v = \setX$. The $|\Nodes|$-tuple of all node features of $\Graph$ is denoted by $\MX = (\vxx_v)_{v \in \Nodes}$. For a set of nodes $\setA \subset \Nodes$, the $|\setA|$-tuple of node features, corresponding to  nodes $\setA$, is denoted by $\featureSubset_{\setA} = (\vxx_v)_{v \in \setA}$.

\subsection{Simple Examples of \Graph-Compatible Functions}
\label{sec:app-inferenceInGraphicalModels}

Compatible functions naturally arise when performing inference on probabilistic graphical models.

\myParagraph{Probabilistic Graphical Models} The joint probability distribution of a probabilistic graphical model, on an undirected graph $\Graph$, is given by
\begin{equation}
	p(\MX~|~\Graph) = \prod_{C \in \CliqueSetOf{\Graph}}\psi_{C}(\vxx_{C}),
\end{equation}
where $\CliqueSetOf{\Graph}$ is the collection of maximal cliques in $\Graph$ and $\psi_{C}$ are some functions, often referred to as \emph{clique potentials}~\cite{Jordan02book, Koller09book}.
This can be written as
\begin{equation}
	p(\MX~|~\Graph) = \exp\{ f(\MX)\},
\end{equation}
where $f(\MX)$ is a \Graph-compatible function according to~\eqref{eq:compatible} with $\theta_{C}(\vxx_C) = \log \psi_{C}(\vxx_C)$. Thus, the ability to approximate any \Graph-compatible function is equivalent to the ability to approximate any distribution function of a probabilistic graphical model, on graph \Graph.

We now provide two examples where we have to learn graph compatible functions to compute maximum likelihood estimates over graphs.

\myParagraph{Graph Classification} Given a graph $\Graph$ and its label $y \in \LabelClass$, suppose that the node features $\MX$ are distributed according to a probabilistic graphical model on the undirected graph $\Graph$. This induces a natural correlation between the observed node features, which is dictated by the graph $\Graph$. Then, the conditional probability density $p(\MX | y, \Graph)$ of the node features $\MX$, given label $y$ and graph $\Graph$, is given by
\begin{equation}
	\label{eq:pgmLG}
	p\left(\MX |y, \Graph \right) = \prod_{C \in \CliqueSetOf{\Graph}} \psi_{C}(x_C, y),
\end{equation}
where $\CliqueSetOf{\Graph}$ is the set of all maximal cliques in graph $\Graph$ and $\psi_{C}$ are the clique potentials. A maximum likelihood estimator for the graph labels will predict:
\begin{equation}
	\label{eq:gc-examp-compatible}
	\hat{y} = \arg\max_{y \in \LabelClass}~\log p\left(\MX |y, \Graph\right) = \arg\max_{y \in \LabelClass} \sum_{C \in \CliqueSetOf{\Graph}} \log \psi_{C}(x_C, y),
\end{equation}
whose objective is a \Graph-compatible function. In practice, we do not know the functions $\psi_{C}$ or the conditional distribution $p(\MX | y, \Graph)$, and will have to learn from the data to make predictions given in~\eqref{eq:gc-examp-compatible} feasible.

\myParagraph{Node Classification} 
%
%
Given a graph $\Graph$ and node labels $\vy = \{ y_{v} \}_{v \in \Nodes(\Graph)}$, suppose the node features $\MX$ be distributed according to a probabilistic graphical model on graph \Graph. We, therefore, have
\begin{equation}
	p( \MX |\vy, \Graph ) = \prod_{C \in \CliqueSetOf{\Graph}} \psi_{C}(\vxx_C, \vy).
\end{equation}

A maximum likelihood estimator that estimates node labels $\vy$ by observing the node features will predict:
\begin{equation}
	\hat{\vy} = \argmax_{y_v \in \LabelClass}~\log p( \MX | \vy, \Graph )
	= \argmax_{y_v \in \LabelClass} \sum_{C \in \CliqueSetOf{\Graph}} \log \psi_{C}(\vxx_C, \vy), 
\end{equation}
whose objective is a \Graph-compatible function. 

\begin{remark}[Applying to directed graphs]
	The proposed model can be used to approximate inference on the directed graphical models as well. Note that the joint distribution on a directed graphical model can also be described as a product of clique potentials~\cite{Jordan02book, Koller09book}. However, we would first convert the directed model into an undirected graphical model using the technique of moralization~\cite{Jordan02book, Koller09book}. The H-tree can then be constructed on this undirected, moralized graph.
\end{remark}

\subsection{Junction Tree Decomposition}
\label{sec:app-junctionTree}

This section reviews the \emph{junction tree} algorithm, proposed in~\cite{Jensen94uai}.
We denote by $(\Tree, \calB) = \texttt{junction-tree}(\Graph)$ the algorithm that takes an
arbitrary graph $\Graph$ and returns a junction tree decomposition $(\Tree, \calB)$ as described below. 

In order to obtain a junction tree decomposition of a given undirected graph $\Graph$, the graph $\Graph$ is first triangulated. Triangulation is done by adding a chord between any two nodes in every cycle of length $4$ or more. This eliminates all the cycles of length $4$ or more in the graph $\Graph$ to produce a chordal graph $\Graph_{c}$.
The collection of bags $\bag = \{ B_{\tau} \}_{\tau}$ in the junction tree is chosen as the set of all maximal cliques in the chordal graph $\Graph_{c}$.
Then, an \emph{intersection graph} $\calI$ on $\bag$ is built, which has a node for every bag in $\bag$ and an edge between two bags $B_{\tau}$ and $B_{\mu}$ if they have a non-empty intersection, \ie $|B_{\tau}\cap B_{\mu}| \geq 1$. The weight of every link $\{\tau, \mu\}$ in the intersection graph $\calI$ is set to $|B_{\tau}\cap B_{\mu}|$. Finally, the desired junction tree is obtained by extracting a maximum weight spanning tree on the weighted intersection graph $\calI$. It is know that this extracted tree $\Tree$, with the bag $\calB$, is a valid tree-decomposition of $\Graph$ that satisifes the connectedness and covering property.

The junction tree decomposition of a graph and its subgraphs is shown in Fig.~\ref{fig:jth-illustrate}.

\subsection{Invariant and Equivariant Function Approximation}
\label{sec:app-invariant-equivariant}

In this section, we prove Theorem~\ref{thm:invariant-equivariant}.
We first recall the definitions of \Graph-invariant and \Graph-equivariant functions.
Let $\MX^{\sigma}$ to denote the tuple $(\vxx_{\sigma(v)})_{v \in \Nodes}$, where $\sigma$ is a permutation of nodes $\Nodes$ in graph \Graph.
Define $\Edges^{\sigma} = \{ (\sigma(u), \sigma(v))~|~(u, v) \in \Edges \}$, for edges \Edges~in graph \Graph, and note that $\Graph^{\sigma} = (\Nodes, \Edges^{\sigma})$ is a permutation of graph \Graph.
%
%
\begin{definition}[\Graph-invariant function]
	A function $h: (\setX, \Graph) \rightarrow \mathbb{R}$ is invariant with respect to graph \Graph or \Graph-invariant if
	\begin{equation}
		h(\MX^{\sigma}, \Graph^{\sigma}) = h(\MX, \Graph),
	\end{equation}
	for all permutations $\sigma$ on $\Nodes(\Graph)$.
\end{definition}

\begin{definition}[\Graph-equivariant function]
	A function $h: (\setX, \Graph) \rightarrow \mathbb{R}^{n}$ is equivariant with respect to graph \Graph or \Graph-equivariant if
	\begin{equation}
		h(\MX^{\sigma}, \Graph^{\sigma}) = h(\MX, \Graph)^{\sigma},
	\end{equation}
	for all permutations $\sigma$ on $\Nodes(\Graph)$, where for a $\vz \in \mathbb{R}^n$, $\vz^{\sigma} \in \mathbb{R}^n$ is such that $\vz^{\sigma}_{i} = \vz_{\sigma(i)}$ for all $i \in [n]$.
\end{definition}

Theorem~\ref{thm:invariant-equivariant} can restated, in more detail, as follows:
%
%
\begin{theorem}
	\label{thm:invariance}
	The following statements hold true.
	\begin{enumerate}
		\item For any continuous \Graph-invariant function $h$ and a scalar $\epsilon > 0$ there exists an integer $M \geq 1$ and a collection of $M$ continuous \Graph-compatible functions $\{ f^{i} \}_{i=1}^{M}$ such that
		\begin{equation}
			\sup_{\MX \in \setX}~~\left| h(\MX, \Graph) - \sum_{i=1}^{M}\phi\left( f^{i}(\MX, \Graph)\right)\right| < \epsilon,
		\end{equation}
		where $\phi : \mathbb{R} \rightarrow \mathbb{R}$ is some function.
		\item For any continuous \Graph-equivariant function $h$ and a scalar $\epsilon > 0$ there exists a set of integers $M_l \geq 1$, for $l \in [n]$, and \Graph-compatible functions $\{ f^{l,i}\}_{i=1}^{M_l}$ such that
		\begin{equation}
			\sup_{\MX \in \setX}~~\left| h_l(\MX, \Graph) - \sum_{i=1}^{M_l}\phi\left( f^{l,i}(\MX, \Graph)\right)\right| < \epsilon,
		\end{equation}
		for all $l \in [n]$, where $h_l(\MX, \Graph) \in \mathbb{R}$ denotes the $l$th component of $h$ and $\phi : \mathbb{R} \rightarrow \mathbb{R}$ is some function.
	\end{enumerate}
\end{theorem}
\begin{IEEEproof}
	The proof is based on a result presented in~\cite{Maehara19-universal}.
	%
	%
	Let $\MW$ denote a $n\times n$ adjacency matrix for graph \Graph (\ie $w(u, v) = 0$ if the link $(u, v) \notin \Edges(\Graph)$) and $w(u, v)$ denotes the $(u, v)$th element in $\MW$. Let $\setW$ denote the space of all such adjacency matrices $\MW$ (for graph \Graph) such that $||\MW||_{\infty} \leq 1$, \ie $|w(u,v)| \leq 1$ for all $u, v \in [n]$.
	Let $\setG$ denote the set of all simple graph, \ie graphs with no self-loops or multi-edges.
	
	For a $\MW \in \setW$ and a graph $\calH \in \setG$ define the function:
	\begin{multline}\label{eq:hom_def}
		\homnum{\calH, \MW} = \sum_{\pi \in \mathbb{M}(\calH, \Graph)}~~\prod_{u \in \Nodes(\calH)} w(\pi(u), \pi(u)) \\
		\times \prod_{(u,v) \in \Edges(\calH)} w(\pi(u), \pi(v)).
	\end{multline}
	where $\mathbb{M}(\calH, \Graph)$ denotes the set of all maps $\pi$ from $\Nodes(\calH)$ to $\Nodes(\Graph)$.
	Further, let $\calA$ denote the following class of functions:
	\begin{equation}\label{eq:hom_fun_class}
		\nonumber
		\calA = \left\{ \MW \rightarrow \sum_{\calH \in \setH} \alpha_{\calH} \homnum{\calH, \MW + 2\MI}~\middle|~\begin{array}{c}
			\alpha_{\calH} \in \mathbb{R} \\
			\setH \subset \setG \\
			\setH~\text{is finite}
		\end{array} \right\},
	\end{equation}
	where $\MI$ denotes the $n\times n$ identity matrix. For a $\MW \in \setW$, graph $\calH \in \setG$, and a node $s \in \Nodes(\Graph)$ define the function $\Homnum{\calH, \MW}{} \in \mathbb{R}^n$ such that its $s$th ($s \in \Nodes(\Graph)$) component is given by
	\begin{multline}\label{eq:HOM_def}
		\Homnum{\calH, \MW}{s} = \sum_{\substack{
				\pi \in \mathbb{M}(\calH, \Graph), \\
				\pi(1) = s}}~\prod_{u \in \Nodes(\calH)} w(\pi(u), \pi(u))\\ 
		\times \prod_{(u,v) \in \Edges(\calH)} w(\pi(u), \pi(v)).
	\end{multline}
	Define the function space:
	\begin{equation}\label{eq:HOM_fun_class}
		\nonumber
		\bar{\calA} = \left\{ \MW \rightarrow \sum_{\calH \in \setH} \alpha_{\calH} \Homnum{\calH, \MW + 2\MI}{}~\middle|~\begin{array}{c}
			\alpha_{\calH} \in \mathbb{R} \\
			\setH \subset \setG \\
			\setH~\text{is finite}
		\end{array} \right\}.
	\end{equation}
	We have the following result from~\cite{Maehara19-universal}.
	
	\begin{theorem}[\!\cite{Maehara19-universal}]
		\label{thm:inv_dense}
		The following statements are true:
		
		\begin{enumerate}
			\item $\calA$ is dense in the space of continuous \Graph-invariant functions.
			\item $\bar{\calA}$ is dense in the space of continuous \Graph-equivariant functions.
		\end{enumerate}
	\end{theorem}
	\begin{IEEEproof}
		The only difference between the spaces $\calA$, $\bar{\calA}$ in~\cite{Maehara19-universal} and defined here is that here we fix the input graph \Graph and restrict the space $\setW$ to be the set of all weighted adjacency matrices of \Graph (with bounded weights). However, the exact same arguments presented in~\cite{Maehara19-universal} hold in this case towards establishing the statements in Theorem~\ref{thm:inv_dense}.
	\end{IEEEproof}
	
	We now show how Theorem~\ref{thm:inv_dense} can be translated to establish Theorem~\ref{thm:invariance}. We only present the arguments here for the \Graph-invariant case in Theorem~\ref{thm:invariance}, and the \Graph-equivariance case can be deduced using the same line of arguments.
	
	Firstly, note that any $\homnum{\calH, \MW}$ (in~\eqref{eq:hom_def}) can be written as:
	\begin{multline}\label{eq:homX_def}
		\homnumX{\calH, \MX} = \sum_{\pi \in \mathbb{I}(\calH, \Graph)}~\prod_{u \in \Nodes(\calH)} \theta_{\pi(u)}(\vxx_{\pi(u)}) \\
		\times \prod_{(u,v) \in \Edges(\calH)} \theta_{\pi(u),\pi(v)}(\vxx_{\pi(u)}, \vxx_{\pi(v)}),
	\end{multline}
	for some input node features $\MX = (\vxx_v)_{v \in \Nodes(\Graph)}$ and functions $\theta_{u}, \theta_{u, v}$ for all $u \in \Nodes(\Graph)$ and $(u, v) \in \Edges(\Graph)$ such that $||\theta_{u}||_{\infty} \leq 1$ and $|| \theta_{u, v}|| \leq 1$ (This follows from $|| \MW ||_{\infty} \leq 1$). Furthermore, the reverse is also true, \ie, for every $\homnumX{\calH, \MX}$ defined in~\eqref{eq:homX_def} there exists a weighted adjacency matrix $\MX$, with $|| \MW ||_{\infty} \leq 1$, such that $\homnumX{\calH, \MX} = \homnum{\calH, \MW}$ (define $w(u, u) = \theta_{u}(\vxx_u)$ and $w(u, v) = \theta_{u, v}(\vxx_u, \vxx_v)$ to get the required $\MW$).

	This observation, in conjunction with Theorem~\ref{thm:inv_dense}, shows that the set of functions
	\begin{equation}\label{eq:homX_fun_class}
		\nonumber
		\calB = \left\{ \MX \rightarrow \sum_{\calH \in \setH} \alpha_{\calH}~\homnumX{\calH, \MX}  \middle| \begin{array}{c}
			\alpha_{\calH} \in \mathbb{R},~\setH \subset_{\text{finite}} \setG \\
			\theta_{u} = \theta_{u}^{'} + 2, \\
			||\theta_u||_{\infty} \leq 1,~\text{and} \\
			|| \theta_{u, v}||_{\infty} \leq 1
		\end{array}  \right\},
	\end{equation}
	is also dense in the space of continuous \Graph-invariant functions. We now show that every function in $\calB$ can be written as a finite sum of \Graph-compatible functions composed with a non-linear function.
	\begin{lemma}
		\label{lem:homX_and_Gcompatible}
		For every $g \in \calB$ there exists a finite set of \Graph-compatible functions $\{ f^{i} \}_{i=1}^{M}$ and a non-linear function $\phi: \mathbb{R} \rightarrow \mathbb{R}$ such that
		\begin{equation}
			\label{eq:lem:homX_and_Gcompatible}
			g(\MX) = \sum_{i=1}^{M}\phi\left( f^{i}(\MX)\right).
		\end{equation}
		Furthermore, $\phi$ are independent of $g \in \calB$.
	\end{lemma}
	\begin{IEEEproof}
		A function $g \in \calB$ is given by
		\begin{multline}\label{eq:aweo}
			g(\MX) = \sum_{\calH \in \setH}~\sum_{\pi \in \mathbb{M}(\calH, \Graph)} \alpha_{\calH} \prod_{u \in \Nodes(\calH)} \theta_{\pi(u)}(\vxx_{\pi(u)}) \\
			\times \prod_{(u,v) \in \Edges(\calH)} \theta_{\pi(u),\pi(v)}(\vxx_{\pi(u)}, \vxx_{\pi(v)}),
		\end{multline}
		for some $\alpha_{\calH}$, $\theta_{u}$, and $\theta_{u, v}$s. Note that the expression
		\begin{equation}
			\prod_{u \in \Nodes(\calH)} \theta_{\pi(u)}(\vxx_{\pi(u)}) 
			\prod_{(u,v) \in \Edges(\calH)} \theta_{\pi(u),\pi(v)}(\vxx_{\pi(u)}, \vxx_{\pi(v)}),
		\end{equation}
		can be written as $\phi(f^{\calH, \pi}(\MX))$ with $\phi(x) = \exp\{x\}$ and $f^{\calH, \pi}(\MX)$ a \Graph-compatible function given by
		\begin{multline}
			f^{\calH, \pi}(\MX) = \sum_{u \in \Nodes(\Graph)} \log \theta_{\pi(u)}(\vxx_{\pi(u)}) \\
			+ \sum_{(u,v) \in \Edges(\Graph)} \log \theta_{\pi(u),\pi(v)}(\vxx_{\pi(u)}, \vxx_{\pi(v)}).
		\end{multline}
		Thus, we have
		\begin{equation}
			g(\MX) = \sum_{\calH \in \setH}\sum_{\pi \in \mathbb{M}(\calH, \Graph)} \phi\left( f^{\calH, \pi}(\MX) \right),
		\end{equation}
		where we have modified $f^{\calH, \pi}$ to incorporate the constant $\alpha_{\calH}$. Since $\setH$ and $\mathbb{M}(\calH, \Graph)$ are finite sets, we have the result.
		%
	\end{IEEEproof}
	
	The result in Theorem~\ref{thm:invariance} follows from Lemma~\ref{lem:homX_and_Gcompatible} and the observation that $\calB$ is dense in the space of  continuous \Graph-invariant functions.

\end{IEEEproof}

\subsection{Proof of Theorem~\ref{thm:approx}}
\label{sec:app-approximation}
\newcommand{\aDAG}{\ensuremath{\vec{\calD}}}

The proof is divided into four sub-sections. Here is a brief outline:

\myParagraph{1} In Section~\ref{sec:aggregation}, we first prove an \emph{aggregation lemma}. It (roughly) states the following: If the representation vectors at the root nodes of the \Htree $\JTH{\Graph}$ are $\{ \vh_{r} \}_{r \in R}$, at some iteration $t$, then in finitely many more message passing iterations it is possible to output a label $y_{v_0} = \sum_{r \in R} \vh_{r}$.

\myParagraph{2} In Section~\ref{sec:factorization}, we then prove that any \Graph-compatible function $f$ can be written as a sum $f(\MX) = \sum_{r \in R} \gamma_r$ of component functions $\gamma_r$.

\myParagraph{3} In Section~\ref{sec:comp_str}, we establish that the component functions $\gamma_r$ have a compositional structure that matches with the sub-tree $\calT_{r}$ of the \Htree $\JTH{\Graph}$ formed by the root node $r$ and its descendants. This helps in efficient computation of the component function $\gamma_r$ on the sub-tree $\calT_{r}$.

\myParagraph{4} The goal is to first estimate each component $\gamma_r$, by message passing on $\calT_{r}$, and then aggregate by applying the aggregation lemma. In Section~\ref{sec:propagation}, we put it all together to argue that it is indeed possible to approximate any (adequately smooth and bounded) compatibility function $f$, to arbitrary precision $\epsilon$, by the message passing described in~\eqref{eq:theory_model_Agg}. We obtain a bound on the number of parameters $N$ required to approximate any such function in Section~\ref{sec:propagation}.

\subsubsection{Aggregation}
\label{sec:aggregation}
Let the $\COMB$ function be a simple average function:
\begin{multline}
	y_{v_0} = \COMB\left(\{\vh^{\NumIterations}_{l}~|~l~\text{leaf node in $\JTH{\Graph}$~s.t.}~\kappa(l) = v_0\}\right) \\
	\triangleq \frac{1}{\left|\{l~|~\kappa(l) = v_0 \}\right|}\sum_{l: \kappa(l) = v_0} \vh^{T}_{l}, \label{eq:app-comb}
\end{multline}
for some $T$, where index $l$ is over the set of leaf nodes in $\JTH{\Graph}$.
We first prove the following lemma.
\begin{lemma}[Aggregation]
	\label{lem:aggregation}
	Let $\vh_{r}^{t}$ denote the representation vectors of root nodes $r \in R$ at some iteration $t$. If $\vh_{r}^{t} \in  [0, 1]$ for all $r \in R$ and $\sum_{r \in R} \vh^{t}_{r} \in [0, 1]$, then there exists $t_0$ message passing iterations such that
	\begin{equation}
		y_{v_0} = \sum_{r \in R} h_{r}^{t},
	\end{equation}
	for $T = t + t_0$. Further, the parameters used in this message passing and the number of iterations $t_0$ do not depend on $\{ \vh_{r}^{t} \}_{r \in R}$.
\end{lemma}

\emph{Proof:} We first make a few assertions about the message passing described in~\eqref{eq:theory_model_Agg}, in the paper. The proof of the lemma directly follows from them. The assertions are self-evident and we only give a one line descriptive proof following its statement.

\myParagraph{Assertion 1} Let $(v, u)$ be an edge in the \Htree $\JTH{\Graph}$. If $\vh^{t-1}_{v} \in [0, 1]$ then there exists parameters $N_u$, $a^{k}_{u, t}$, $b^{k}_{u, t}$, and $\vw^{k}_{u,t}$ in ~\eqref{eq:theory_model_Agg} such that
\begin{multline}
	\vh^{t}_{u} = \AGG_t\left(\vh^{t-1}_{u}, \{ \vh^{t-1}_{w}~|~w \in \nbr{u}{\JTH{\Graph}} \}\right) \\
	= \ReLU{\sum_{k=1}^{N_u} a_{u, t}^{k}\inner{\vw_{u, t}^{k}}{\vh^{t-1}_{\bar{\calN}(u)}} + b_{u, t}^{k} }, \\
	= \ReLU{\vh^{t-1}_{v}} = \vh^{t-1}_{v}.
\end{multline}
\emph{The last equality holds only because $\vh^{t-1}_{v} \in [0, 1]$.}

\myParagraph{Assertion 2} Let $(v, u)$ be an edge in $\JTH{\Graph}$. If $\vh^{t-1}_{u} + \vh^{t-1}_{v} \in [0, 1]$ then there exists parameters $N_u$, $a^{k}_{u, t}$, $b^{k}_{u, t}$, and $\vw^{k}_{u,t}$ in ~\eqref{eq:theory_model_Agg} such that
\begin{multline}
	\vh^{t}_{u} = \AGG_t\left(\vh^{t-1}_{u}, \{ \vh^{t-1}_{w}~|~w \in \nbr{u}{\JTH{\Graph}} \}\right) \\
	= \ReLU{\sum_{k=1}^{N_u} a_{u, t}^{k}\inner{\vw_{u, t}^{k}}{\vh^{t-1}_{\bar{\calN}(u)}} + b_{u, t}^{k} }, \\
	= \ReLU{ \vh^{t-1}_{u} + \vh^{t-1}_{v}} = \vh^{t-1}_{u} + \vh^{t-1}_{v}.
\end{multline}
\emph{The last equality holds only because $\vh^{t-1}_{u} + \vh^{t-1}_{v} \in [0, 1]$.}

\myParagraph{Assertion 3} Let $\vh^{t}_{r}$ denote representation vectors at root nodes $r \in R$ on the \Htree $\JTH{\Graph}$ at some iteration $t$. If $\vh^{t}_{r} \in [0, 1]$ and $\sum_{r \in R} \vh^{t}_{r} \in [0, 1]$ then for any $r_0 \in R$ there exists $t_0$ message passing iterations, for some $t_0 > 0$, on the root nodes in $\JTH{\Graph}$ such that  $\vh^{t+t_0}_{r_0} = \sum_{r \in R} \vh^{t}_{r}$. Further, the parameters used in this message passing are independent of $\{ \vh^{t}_{r} \}_{r \in R}$.

\emph{This can be established by looking at $\Tree = \JTH{\Graph}[R]$ as a tree rooted at $r_0$ and performing message aggregation from the leaf nodes of $\Tree$ to the root node $r_0$ using Assertion~2.}

\myParagraph{Assertion 4} If $\vh^{t}_{r} \in [0, 1]$ for some $r \in R$, then there exists $t_0$ message passing iterations from the root node $r$ to all the leaf nodes $l$ in \Htree $\JTH{\Graph}$ such that $\vh^{t+t_0}_l = \vh^{t}_{r}$, for all leaf nodes $l$.

\emph{This can be done by using Assertion~1 and successively passing the representation vector $\vh^{t}_r$ from $r$ to all the leaf nodes $l$ in $\JTH{\Graph}$.}

From Assertions~3 and~4 it is clear that, given $\{ \vh^{t}_{r} \}_{r \in R}$ at some $t$ (bounded in $[0, 1]$ as described in the statement of the lemma), there exists $t_0$ message passing iterations such that $\vh^{t+t_0}_{l} = \sum_{r \in R} \vh^{t}_{r}$ at all the leaf nodes $l$ in $\JTH{\Graph}$. Since the \COMB~operation computes a simple average (see~\eqref{eq:app-comb}) we have the result.~\qed

\subsubsection{Factorization}
\label{sec:factorization}
Next, we show that any compatibility function
\begin{equation}
	\label{eq:app_compattible_fun}
	f(\MX) = \sum_{C \in \CliqueSetOf{\Graph}} \theta_{C}(\vxx_{C}),
\end{equation}
can be broken down into component functions $\{ \gamma_{r} \}_{r \in R}$ such that
\begin{equation}
	\label{eq:compat_local}
	f(\MX) = \sum_{r \in R} \gamma_r,
\end{equation}
where
\begin{equation}
	\label{eq:def_gamma}
	\gamma_r = \sum_{C \in \calC_{r}} \theta_{C}(\vxx_{C}),
\end{equation}
for all $r \in R$,\footnote{We omit the explicit dependence of the function $\gamma_r$ on $\MX$ to ease the notation.} $\calC_{r}$ are subsets of $\CliqueSetOf{\Graph}$ which form its partition, and $R$ is the set of root nodes in the \Htree $\JTH{\Graph}$.
%
%
\begin{lemma}[Factorization]
	\label{lem:factorization}
	Let $f$ be a graph compatible function given in~\eqref{eq:app_compattible_fun} with its clique functions $\theta_{C}$. Then, for every $r \in R$ there exists a subset $\calC_{r} \subset \CliqueSetOf{\Graph}$ such that
	\begin{equation}
		\gamma_r = \sum_{C \in \calC_{r}} \theta_{C}(x_C),
	\end{equation}
	and $f(\MX) = \sum_{r \in R} \gamma_r$. Further, the collection of subsets $\{ \calC_{r} \}_{r \in R}$ forms a partition of $\CliqueSetOf{\Graph}$, \ie $\calC_{r}\cap \calC_{r'} = \emptyset$ whenever $r \neq r'$ and $\cup_{r \in R}\calC_{r} = \CliqueSetOf{\Graph}$.
\end{lemma}

\emph{Proof:} Let $f$ be a graph compatible function given in~\eqref{eq:app_compattible_fun} with its clique functions $\theta_{C}$ and
\begin{equation}
	(\Tree, \calB) = \texttt{tree-decomposition}(\Graph),
\end{equation}
be the \treeDecomposition of graph $\Graph$. Note that the set of root nodes $R$, in the \Htree $\JTH{\Graph}$, is in fact all the nodes in $\Tree$, namely $R = \Nodes(\Tree)$. Further, for every $r \in R$, $B_{r} \in \calB$ is a bag of nodes $B_{r} \subset \Nodes(\Graph)$ associated with $r$.

It is known that for any clique $C$ in graph $\Graph$, \ie $C \in \CliqueSetOf{\Graph}$, there exists an $r \in R$ such that all nodes in $C$ are in the bag $B_{r}$, \ie $\Nodes(C) \subset B_{r}$~\cite{Diestel05book_graphTheory}.
However, it is possible that two bags $B_{r}$ and $B_{r'}$, for $r \neq r'$, may contain all the nodes of the same clique $C$.

Ideally, we would define
\begin{equation}
	\calC_{r} \triangleq \left\{ C \in \CliqueSetOf{\Graph}~|~\Nodes(C) \subset B_{r}\right\},
\end{equation}
which is the set of all cliques $C$ in $\Graph$ such that all its nodes are in the bag $B_{r}$,
and the functions $\gamma_r$ to be
\begin{equation}
	\gamma_r = \sum_{C \in \calC_{r}} \theta_{C}(\vxx_{C}),
\end{equation}
for all $r \in R$. However, this can lead the $\sum_{r \in R} \gamma_r$ to overestimate the function $f$. This is because two bags $B_{r}$ and $B_{r'}$ may contain all the nodes of the same clique $C$.

In order to avoid double counting of clique functions, we order the nodes in $R$ as $R = \{r_1, r_2, \ldots r_{|R|} \}$. We then iterate over these ordered $R$ nodes in the tree-decomposition to generate $\calC_{r_k}$ and $\gamma_{r_k}$ (for $k = 1, 2, \ldots |R|$) as follows. Initialize $\calM_{1} = \emptyset$ and iterate over $k = 1, 2, \ldots |R|$:
\begin{equation}
	\label{eq:nut1}
	\calC_{r_k} = \left\{ C \in \CliqueSetOf{\Graph}\backslash \calM_{k}~|~\Nodes(C) \subset B_{r_k}\right\},
\end{equation}
\begin{equation}
	\label{eq:nut2}
	\calM_{k+1} = \calM_{k}\cup \calC_{r_k},
\end{equation}
and set
\begin{equation}
	\gamma_{r_k} = \sum_{C \in \calC_{r_{k}}} \theta_{C}(\vxx_{C}),
\end{equation}
for $k = 1, 2, \ldots |R|$. This procedure ensures that we do not overestimate $f$ and have $f(\MX) = \sum_{r \in R} \gamma_{r}$.

Furthermore, $\{ \calC_{r} \}_{r \in R}$ by its very construction (in~\eqref{eq:nut1}-\eqref{eq:nut2}) is pairwise disjoint and spans the entire $\CliqueSetOf{\Graph}$, thereby forming its partition. ~\qed



\subsubsection{Compositional Structure}
\label{sec:comp_str}
\begin{figure}
	\centering
	\includegraphics[width=0.85\linewidth]{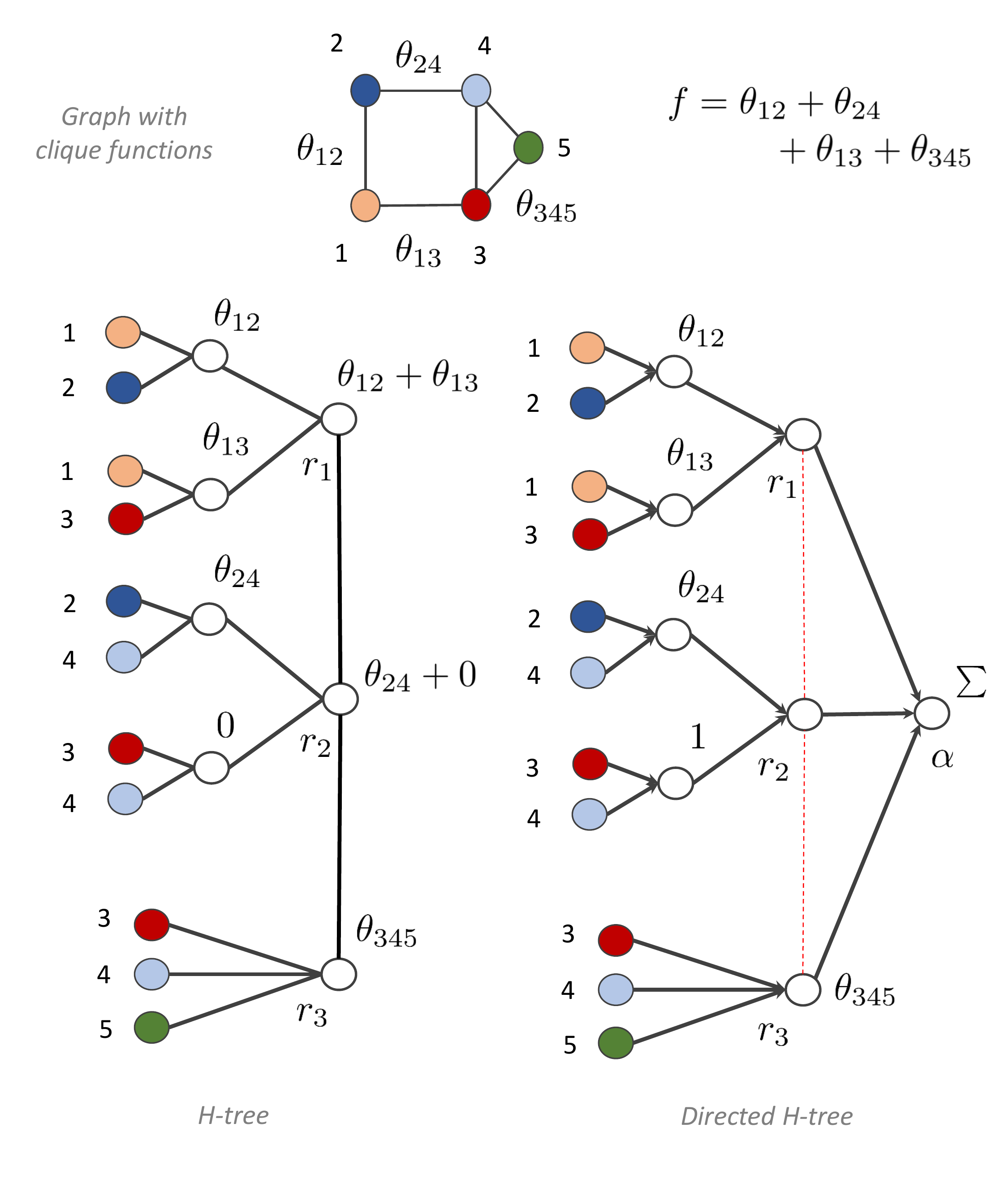}
	\caption{Shows the \Htree $\JTH{\Graph}$ and the directed \Htree $\dJTH{\Graph}$ of a graph. Computation of a compatible function $f$ is shown on the \Htree.}\label{fig:jth-fun-compute}
\end{figure}

Fig.~\ref{fig:jth-fun-compute} illustrates computation of a compatible function on the \Htree. We see how the computation of $f$ splits as $f = \gamma_{r_1} + \gamma_{r_2} + \gamma_{r_3}$, where $\gamma_{r_1} = \theta_{12} + \theta_{13}$, $\gamma_{r_2} = \theta_{24}$, and $\gamma_{r_3} = \theta_{345}$. It is interesting to note that the functions $\gamma_{r}$, further, have a compositional structure that matches with the sub-tree induced by the root nodes $r$, and its descendants. For example, the compositional structure of $\gamma_{r_1}$ matches with the sub-tree formed by the root node $r_1$ and its descendants in the \Htree $\JTH{\Graph}$. This turns out to be true in general for any compatible function $f$, and its factorization $\{ \gamma_r \}_{r \in R}$ (in Lemma~\ref{lem:factorization}).

In order to make this precise, we introduce a few definitions that are inspired by~\cite{Poggio2017a-arXiv-WhyWhenDNN, Poggio2017j-IJAC-WhyWhenDNN}.
Let $\aDAG$ be a directed acyclic graph (DAG) with a single root node $\alpha$, i.e. all the directed paths in $\aDAG$ end at $\alpha$. We will use the term DAG to refer to a single-root DAG in this section.
\begin{definition}
	\label{def:comp-fun_matching}
	A function $f: \MX = (x_i)_{i \in [n]} \rightarrow f(\MX) \in \mathbb{R}$ is said to have a compositional structure that matches with a DAG $\aDAG$, with root node $\alpha$, if the following holds:
	
	1. Each leaf node $l$ of $\aDAG$ embeds one component of the input feature, \ie $\vh_l = \vxx_i$ for some $i \in [n]$.
	
	2. For every non-leaf node $u$ there exists some function $\calH_{u}$ such that
	\begin{equation}
		\vh_{u} = \calH_{u}\left( \{ \vh_{w}~|~w \in \calN_{\text{in}}(u)\}\right),
	\end{equation}
	where $\calN_{\text{in}}(u)$ denotes the set of all incoming neighbors to node $u$.
	
	3. $f(\MX) = \vh_{\alpha}$, where $\alpha$ is the single-root node of $\aDAG$.
\end{definition}


Let $\calT_{r}$ denote the sub-tree of $\JTH{\Graph}$ induced by the root $r \in R$ and its descendants in $\JTH{\Graph}$. Further, let $\vec{\calT}_{r}$ denote a directed version of $\calT_{r}$ in which every edge $e$ in $\calT_{r}$ is turned into a directed edge, pointing in the direction of the root node $r$. Note that $\vec{\calT}_{r}$ is a DAG with node $r$ functioning as the single root node.

We now show that the components $\{ \gamma_r \}_{r \in R}$ of the compatible function $f$ in Lemma~\ref{lem:factorization} have a compositional structure that matches with $\vec{\calT}_{r}$.
%
%
\begin{lemma}[Compositional Structure]
	\label{lem:comp_str}
	The function $\gamma_r$, in Lemma~\ref{lem:factorization}, has a compositional structure that matches with the directed sub-tree $\vec{\calT}_{r}$, for all $r \in R$.
\end{lemma}
\emph{Proof:} In Lemma~\ref{lem:factorization}, the function $\gamma_r$ is given by
\begin{equation}
	\gamma_r = \sum_{C \in \calC_{r}} \theta_{C}(\vxx_{C}),
\end{equation}
were $\calC_r$ is given by
\begin{equation}
	\calC_{r} = \left\{ C \in \CliqueSetOf{\Graph}\backslash \calM~|~\Nodes(C) \subset B_{r}\right\},
\end{equation}
for some set $\calM \subset \CliqueSetOf{\Graph}$. The set $\calC_{r}$ can be thought of as a collection of cliques in the subgraph of $\Graph$ induced by the bag $B_{r}$, namely $\Graph[B_{r}]$. Therefore, the function $\gamma_r$ is a compatible function on $\Graph[B_{r}]$.

Note that, in Lemma~\ref{lem:factorization}, we showed that a graph $\Graph$ compatible function can be factored as a sum of $R$ functions, call them $\{ \gamma_r \}_{r \in R}$, where $R$ is the set of nodes in the tree-decomposition $(\Tree, \calB)$. We have now argued that the functions $\gamma_r$ are compatible function on the subgraphs $\Graph[B_r]$.

Note that the set of all children of node $r$ in the sub-tree $\calT_{r}$ (of the \Htree $\JTH{\Graph}$) form a tree decomposition of $\Graph[B_r]$. This indicates that the function $\gamma_r$ should also split as a sum of functions, one corresponding to each node in the tree decomposition of $\Graph[B_r]$, by Lemma~\ref{lem:factorization}.

Thus, by successively applying Lemma~\ref{lem:factorization}, we can see that the compositional structure of $\gamma_r$ matches with the directed sub-tree $\vec{\calT}_{r}$, constructed out of the \Htree $\JTH{\Graph}$.~\qed

\subsubsection{Approximation}
\label{sec:propagation}
Lemmas~\ref{lem:aggregation} and~\ref{lem:factorization} suggest that in order to approximate a compatible function $f(\MX) = \sum_{C \in \CliqueSetOf{\Graph}} \theta_{C}(\vxx_C)$, with $f$ and $\theta_{C}$ bounded between $[0, 1]$, it suffices to generate representation vectors
\begin{equation}
	\vh^{t}_{r} \approx \gamma_r = \sum_{C \in \calC_{r}} \theta_{C}(\vxx_C), \label{eq:here1}
\end{equation}
at each root node $r \in R$ of the \Htree $\JTH{\Graph}$, for some $t$. The approximation in~\eqref{eq:here1} must be such that
\begin{equation}
	\left| \sum_{r \in R}\vh^{t}_{r} - \sum_{r \in R}\gamma_r \right| = \left| \sum_{r \in R}\vh^{t}_{r} - f(\MX) \right| < \epsilon. \label{eq:here2}
\end{equation}
Once such representation vectors $\vh^{t}_{r}$ are generated at the root nodes of the \Htree, by Lemma~\ref{lem:aggregation}, it's sum can be propagated to generate the node label $y_{v_0} = \sum_{r \in R} \vh^{t}_{r}$, with message passing that is independent of the function being approximated.

Next, we show that the message passing defined in~\eqref{eq:theory_model_Agg} can indeed produce an approximation, give in~\eqref{eq:here2}. The number of parameters required to attain this approximation will be an upper-bound on $N$.

To prove this, we consider a directed version of the \Htree $\JTH{\Graph}$, where each edge in $\JTH{\Graph}$ is turned into a directed edge pointing in the direction that leads to the root nodes $\Root \in \JTH{\Graph}$.
We also remove the edges between the root nodes $\Root$, and add another final node that aggregates information from all the root nodes. We call the final node the \emph{aggregator} and call it $\alpha$.
%
%
We call this directed graph $\dJTH{\Graph}$. A directed \Htree graph is illustrated in Figure~\ref{fig:jth-fun-compute}. The red colored edges between root nodes show the deleted edges between the root nodes $\Root$ in $\JTH{\Graph}$ to get $\dJTH{\Graph}$.

We assume that the messages propagate only in one direction, i.e. from the leaf nodes, where the input node features are embedded, to the aggregator node $\alpha$. We implement a shallow neural network at every non leaf node in $\dJTH{\Graph}$, which takes in input from all its incoming edges, and propagates its output through its single outgoing edge, directed towards the root nodes.

This can be implemented in the original message passing~\eqref{eq:theory_model_Agg} by setting the weight (\ie parameter $w^{k}_{u, t}$) component corresponding to the parent node, in the directed $\JTH{\Graph}$, to zero. This final aggregation layer in $\dJTH{\Graph}$ is only for mathematical purpose so that we can prove an $\epsilon$ approximation result, as in~\eqref{eq:here2}.

With this, in the new message passing architecture on $\dJTH{\Graph}$, each non-leaf node $u$ in $\dJTH{\Graph}$ implements the following shallow neural network given by
\begin{equation}
	\label{eq:shallow_layer}
	\vh_{u} = \ReLU{\sum_{k=1}^{N_u} a_{u}^{k}\inner{\vw_{u}^{k}}{\vh_{\calN_{\text{in}}(u)}} + b_{u}^{k} },
\end{equation}
where $\vh_{\calN_{\text{in}}(u)} \triangleq (\vh_{u'}~|~u' \in \calN_{\text{in}}(u))$ denotes the vector formed by concatenating all the representation vectors $\vh_{u'}$ of nodes $u'$ that have an incoming edge to $u$ in $\dJTH{\Graph}$. Here, $a^{k}_u$ and $b^{k}_u$ are constants and $\vw^{k}_u$ is a vector of size $d_u-1$, which is the total number of incoming links to node $u$ in $\dJTH{\Graph}$ and $d_u$ is the total number of links that node $u$ has in $\JTH{\Graph}$.
Thus, for every non-leaf node $u \in \dJTH{\Graph}$ we have $(d_u - 1)\times N_u$ parameters that model the shallow network.
The aggregator node generates the output by simply summing the representation vectors at the root nodes.

Note that, in~\eqref{eq:shallow_layer}, $\vh_{u}$ depends on the input node features $\MX$. We omit this dependence in the notation for ease of presentation. We now define the space of functions that the above message passing on $\dJTH{\Graph}$ produces:
\begin{equation}
	\label{eq:sapce_djth}
	\calF(\Graph, N) = \left\{ \MX \rightarrow \sum_{u \in \Root} \vh_{u}~\Big|~\vh_{u}~\text{given in~\eqref{eq:shallow_layer}}~\right\},
\end{equation}
where $N = \sum_{u} (d_u - 1) N_{u}$ is the sum of all the parameters used in~\eqref{eq:shallow_layer}.

In the following, we will restrict ourselves to the DAG $\dJTH{\Graph}$ and argue that any (smooth enough) function $f$ that has a compositional structure that matches with $\dJTH{\Graph}$ can be approximated by a $g \in \calF(\Graph, N)$ (see~\eqref{eq:sapce_djth}) with an arbitrary precision.

We now show that for any (smooth enough) function $f$, which has a compositional structure that matches with the directed \Htree $\dJTH{\Graph}$, can be approximated by a $g \in \calF(\Graph, N)$ (see~\eqref{eq:sapce_djth}) with an arbitrary precision.
\begin{theorem}
	\label{thm:poggio}
	Let $f:[0, 1]^{n} \rightarrow [0, 1]$ be a function that has a compositional structure that matches with the DAG $\dJTH{\Graph}$. Let every constituent function $\calH_{u}$ of $f$ (see Definition~\ref{def:comp-fun_matching}) be $L_{u}$-Lipschitz with respect to the infinity norm. Then, for every $\epsilon > 0$ there exists a neural network $g \in \calF(\Graph, N)$ such that $||f - g||_{\infty} < \epsilon$ and the number of parameters $N$ is bounded by
	\begin{equation}
		\label{eq:temp_bound}
		N = \calO\left( \sum_{u \in \Nodes(\dJTH{\Graph})\backslash \{ \alpha \}} (d_{u} - 1) \left( \frac{\epsilon}{L_{u}} \right)^{-(d_u - 1)} \right),
	\end{equation}
	where $d_{u}$ denotes the degree (counting incoming and outgoing edges) for node $u$ in $\dJTH{\Graph}$
\end{theorem}
\emph{Proof:} The proof of this result follows directly from the arguments presented for Theorem 3, Theorem 4, and Proposition 6 in~\cite{Poggio2017a-arXiv-WhyWhenDNN, Poggio2017j-IJAC-WhyWhenDNN}. The first modification we make is the constant factor term $(d_{u} - 1)$ for each node $u$ in the summation in~\eqref{eq:temp_bound}. This appears here, but not in~\cite{Poggio2017a-arXiv-WhyWhenDNN, Poggio2017j-IJAC-WhyWhenDNN}, because in~\cite{Poggio2017a-arXiv-WhyWhenDNN, Poggio2017j-IJAC-WhyWhenDNN} the node degree was considered as a constant. Here, the degree relates to the treewidth of the graph, and is an important parameter to track scalability of the architecture. The second modification is that we allow for different Lipschitz constants $L_{u}$ for different constituent function. However, the arguments in~\cite{Poggio2017a-arXiv-WhyWhenDNN, Poggio2017j-IJAC-WhyWhenDNN} work for this case as well.~\qed

We now apply Theorem~\ref{thm:poggio} to the function $f$ given in the statement of Theorem~\ref{thm:approx}. In it, $f$ is compatible with respect to $\Graph$. Thus, using
Lemma~\ref{lem:factorization} and Lemma~\ref{lem:comp_str}, we can deduce that $f$ also has a compositional structure that matches with the directed \Htree $\dJTH{\Graph}$. In Figure~\ref{fig:jth-fun-compute}, we illustrate this for a simple example. Thus we can apply Theorem~\ref{thm:poggio} on $f$ in order to seek an approximation $g \in \calF(\Graph, N)$.

In applying Theorem~\ref{thm:poggio}, we see that the functions $\theta_C$ are $1$-Lipschitz. Thus, all the nodes $u \in \dJTH{\Graph}$ at which we compute $\theta_C$, $L_{u} = 1 \leq d_u - 1$.
The remaining functions that are to be approximated on the $\dJTH{\Graph}$ are the addition functions (see Figure~\ref{fig:jth-fun-compute} to know how they arise in computing a compatible function). In order to derive our result, it suffices to argue that a simple sum of $k$ variables, taking values in the unit cube $[0, 1]^{k}$, is $k$-Lipschitz with respect to the sup norm. This is indeed true and can be verified by simple arguments in analysis. Thus, for all the nodes $u$ on which we have to compute the addition, we have $L_u = d_{u} - 1$, where $d_{u}$ is the degree of node $u$ (counting both incoming and outgoing edges).

Putting all this together and applying Theorem~\ref{thm:poggio} we obtain the result.

\subsection{Proof of Corollary~\ref{cor:approx}}
\label{sec:app-bound}
We first obtain upper-bounds on the number of nodes $|\Nodes(\JTH{\Graph})|$ in the \Htree $\JTH{\Graph}$ and the node degree $d_u$ for $u \in \Nodes(\JTH{\Graph})$. We prove the desired result by substituting these bounds in Theorem~\ref{thm:approx}.

First, note that the subgraph of the \Htree $\JTH{\Graph}$ induced by the set of root nodes $R$ is a \treeDecomposition $(\Tree = \JTH{\Graph}[R], \calB)$ of $\Graph$, by construction; see Algorithm~\ref{algo:jth} (lines 1-3).
Let $\treewidth{\JTH{\Graph}}$ denote the treewidth of the \treeDecomposition $(\Tree = \JTH{\Graph}[R], \calB)$.
Then the size of each bag $B_{\tau} \in \calB$ is bounded by the treewidth $\treewidth{\JTH{\Graph}} + 1$ (see~\eqref{eq:def_treewidth}).
Let $T_{\tau}$ denote the sub-tree in $\JTH{\Graph}$ that is formed by all the descendants of, and including, the node $\tau$ in $\Tree = \JTH{\Graph}[R]$. Then, the number of nodes in $\JTH{\Graph}$ is given by
\begin{equation}\label{eq:local1}
	|\Nodes(\JTH{\Graph})| = \sum_{\tau \in R} |\Nodes(T_{\tau})|.
\end{equation}

Note that the size of each sub-tree $|\Nodes(T_{\tau})|$ is bounded by
\begin{equation}\label{eq:local2}
	|\Nodes(T_{\tau})| \leq 1 + \left( \treewidth{\JTH{\Graph}} + 1 \right)^{\treewidth{\JTH{\Graph}} + 1}.
\end{equation}
This is because the depth of the tree $T_{\tau}$ is bounded by the bag size $|B_{\tau}|$, which is upper-bounded by $\treewidth{\JTH{\Graph}} + 1$. Further, no node in $T_{\tau}$ has a bag size larger than $|B_{\tau}|$ and therefore the number of children at each non-leaf node in $T_{\tau}$ is bounded by $\treewidth{\JTH{\Graph}} + 1$. The additional ``$+1$'' in~\eqref{eq:local2} accounts for the root node $\tau$ in $T_{\tau}$.

Finally, the number of nodes in the \treeDecomposition $\Tree = \JTH{\Graph}[R]$ (or equivalently, the number of root nodes $R$) is upper-bounded by $n$, the total number of nodes in graph $\Graph$. This, along with~\eqref{eq:local1}-\eqref{eq:local2}, imply
\begin{equation}\label{eq:local3}
	|\Nodes(\JTH{\Graph})| \leq n + n\left( \treewidth{\JTH{\Graph}} + 1\right)^{\treewidth{\JTH{\Graph}}+1}
\end{equation}

Note that the degree minus 1, $d_u - 1$, is the size of the bag in a \treeDecomposition of some subgraph of $\Graph$. Since the size of the largest bag in the \treeDecomposition of the entire graph is bounded by $\treewidth{\JTH{\Graph}} + 1$, we have
\begin{equation}
	\label{eq:degree_bound}
	d_u - 1 \leq \treewidth{\JTH{\Graph}} + 1,
\end{equation}
for all $u \in \Nodes(\JTH{\Graph})$.


Substituting~\eqref{eq:degree_bound}-\eqref{eq:local3} in~\eqref{eq:param_approx} of Theorem~\ref{thm:approx} we obtain the result.



\subsection{Addendum to 3D Scene Graph Experiments}
\label{sec:app-experiments}
We provide more details on the (i)~approaches and setup, (ii)~the compute, train and test time requirements, (iii)~the methods we use for tuning of our hyper-parameters, and (iv)~the list of semantic labels in the dataset.

\myParagraph{Approaches and Setup}
\begin{table}[t]
	\caption{Time Requirements: Train, and Test}
	\label{tab:scene_graph_timing_extra}
	\vspace{-0.1cm}
	\begin{center}
		\begin{small}
			\begin{tabular}{lccccr}
				\toprule
				Model        	    & Training (per epoch)   & Testing 	\\
				\midrule
				GCN                 & 0.072 s	             & 0.048 s  \\
				NT $+$ GCN          & 0.305 s	             & 0.058 s  \\
				GraphSAGE  		    & 0.068 s	             & 0.042 s	\\
				NT $+$ GraphSAGE    & 0.311 s	             & 0.060 s	\\
				GAT       		    & 0.089 s	             & 0.049 s	\\
				NT $+$ GAT         	& 0.872 s	             & 0.107 s	\\
				GIN  		        & 0.079 s	             & 0.043 s	\\
				NT $+$ GIN         	& 0.348 s	             & 0.059 s	\\
				\bottomrule
			\end{tabular}
		\end{small}
	\end{center}
	\vspace{-0.4cm}
\end{table}
We implement the neural tree architecture with
four different aggregation functions $\text{AGG}_t$ specified in: GCN~\cite{Kipf17iclr-gcn}, GraphSAGE~\cite{Hamilton17nips-GraphSage}, GAT~\cite{Velickovic18iclr-GAT}, GIN~\cite{Xu19iclr-gin}. We randomly select 10\% of the nodes for validation and 20\% for testing. 
The hyper-parameters of the two approaches are separately tuned based on the best validation accuracy, while using all 70\% of the remaining nodes for training.
%
%
%
The $\text{READ}$ function for the standard GNN (see~\eqref{eq:GNN_READ}) is implemented as a single linear layer followed by a softmax.
On the other hand, the $\text{COMB}$ function (see~\eqref{eq:comb}) for neural trees is implemented as a mean pooling operation, followed by a single linear layer and a softmax. We use different READ (resp. COMB) functions for the room nodes and the object nodes.
We use the ReLU activation function and also implement dropout at each iteration.
We train the architectures using the standard cross entropy loss function. The experiments are implemented using the PyTorch Geometric library.

\textbf{Time Requirements.} We study the time required for computing, training, and testing our model over the 3D scene graph dataset. It takes about $2.08$ sec to compute \Htrees for all the 482 room-object scene graphs.

In Table~\ref{tab:scene_graph_timing_extra},
we report the train and test time for the standard GNN architectures -- GCN, GraphSAGE, GAT, GIN -- and the corresponding neural trees.
We observe that the neural tree takes about 4x-10x more time to train compared to the corresponding standard GNN. This is expected because the \Htree is much larger than the input graph, and as a consequence, the neural tree architecture needs to train more weights than a standard GNN.
The testing time for the neural trees, on the other hand, remains comparable to the standard GNN architectures. This makes the more accurate neural trees architecture amenable for real-time deployment.
The reported times are measured when implementing the respective models on an Nvidia Quadro P4000 GPU processor.

\myParagraph{Hyper-parameter Tuning}
%
We tune the hyper-parameters in the following order, as recommended by \cite{Shchur2019a-arXiv-GNN-EvalPitfalls}:
\begin{itemize}
	\item Iterations: [1, 2, 3, 4, 5, 6]
	\item Hidden dimension: [16, 32, 64, 128, 256]
	\item Learning rate: [0.0005, 0.001, 0.005, 0.01]
	\item Dropout probability: [0.25, 0.5, 0.75]
	\item $L_2$ regularization strength: [0, 1e-4, 1e-3, 1e-2]
\end{itemize}
We first tune the number of iterations, hidden dimension, and learning rate using a grid search, while keeping dropout and $L_2$ regularization to the lowest value.
For both standard GNN and neural tree, a single choice of the triplet: number of iterations, hidden dimension, and learning rate, yields significantly higher accuracy than the others. With this triplet fixed,
we then tune the dropout and $L_2$ regularization using another grid search.

In training, we notice that the batch size does not have a noticeable impact on the training and test accuracy. After having experimented with various batch sizes between $32$ to $512$, we recommend and use a batch size of $128$ in our experiments.
%


Table~\ref{tab:sg_results} (in Section~\ref{sec:experiments}) reported the test accuracies for various standard GNNs and neural tree models.
The tuned hyper-parameters for these models are given in Table~\ref{tab:sg_hyperparams}. These hyper-parameters were tuned using the procedure described in the previous paragraph. A dropout ratio of 0.25 turns out to be the optimal choice in all cases. The optimization is run for no more than 1000 epochs of SGD (using the Adam optimizer) to achieve reasonable convergence during training.

\begin{table}[b]
	\centering
	\vspace{-0.3cm}
	\caption{Tuned Hyper-parameters for Various Models \label{tab:sg_hyperparams}}
	\scalebox{0.95}{
		\begin{tabular}{lcccc}
			\toprule
			Model          & hidden dim. & iter. & regularization & learning rate \\
			\midrule
			GCN    	            & $64$   & 3	   &0.0 	    & 0.01	\\
			NN + GCN         	& $128$  & 4	   &0.0 	    & 0.01	\\					
			GraphSAGE           & $128$  & 3	   &1e-3 	    & 0.005	\\			
			NN + GraphSAGE   	& $128$  & 4       &1e-3  	    & 0.005	\\				
			GAT                 & $128$  & 2	   &1e-4 	    & 0.001	\\			
			NN + GAT   			& $128$  & 4	   &1e-4 	    & 0.0005	\\
			GIN                	& $64$   & 3	   &1e-3 	    & 0.005	\\			
			NN + GIN  			& $128$  & 4	   &1e-3 	    & 0.005	\\
			\bottomrule
	\end{tabular}}
\end{table}

Apart from the four listed hyper-parameters (hidden dimension, number of iterations, $L_2$ regularization, learning rate), some of the implemented architectures (GAT, GraphSAGE, GIN) have their specific design choices and hyper-parameters.
In the case of GAT, for example, we use 6 attention heads and ELU activation function (instead of ReLU) to be consistent with the original paper.
For GraphSAGE (in Table~\ref{tab:sg_hyperparams}), we use the GraphSAGE-mean from the original paper, which does mean pooling after each convolution operation. 
In the case of GIN, we use the more general GIN-$\epsilon$ and train $\epsilon$ for better performance. 

\myParagraph{Semantic Labels in the Dataset}
In the 482 room-object scene graphs we used for testing, the room labels are: bathroom, bedroom, corridor, dining\_room, home\_office, kitchen, living\_room, storage\_room, utility\_room, lobby, playroom, staircase, closet, gym, garage.
The object labels are: bottle, toilet, sink, plant, vase, chair, bed, tv, skateboard, couch, dining\_table, handbag, keyboard, book, clock, microwave, oven, cup, bowl, refrigerator, cell\_phone, laptop, bench, sports\_ball, backpack, tie, suitcase, wine\_glass, toaster, apple, knife, teddy\_bear, remote, orange, bicycle.

%

\subsection{Addendum to Citation Network Experiments}
\label{sec:app-experiments-cite}

We provide more details on the (i)~hyper-parameter tuning and (ii)~the compute, train and test time requirements.

\textbf{Hyper-parameter Tuning.} We use the same hyper-parameters (hidden dimension, number iterations, number of attention heads) for the neural trees as reported in the original GCN and GAT papers, except the learning rate, $L_2$ regularization, and dropout. These hyper-parameters pertain to the optimization algorithm used for training and are tuned to achieve the best results, \ie highest validation accuracy while not over-fitting. Better performance can be achieved using a specifically tailored message passing function for the neural trees, but the goal here is to understand when message passing on \Htree, \ie neural tree, performs better than message passing on the input graph, \ie standard GNN.

For each dataset, we randomly select $500$ nodes for validation and $1000$ nodes for testing. We vary the training data from $20$ nodes per label, to all the remaining nodes (not used for validation and testing) in the network.
We report the accuracy (and its variance) over $10$ runs. The experiments are performed using PyTorch Geometric.

\textbf{Time Requirements.} We study the time required to compute, train, and test our model over these large citation 
\begin{wrapfigure}[15]{r}{0.6\linewidth}
	\vspace{-0.3cm}
	\centering
	\includegraphics[trim=45 175 30 175,clip,width=\linewidth]{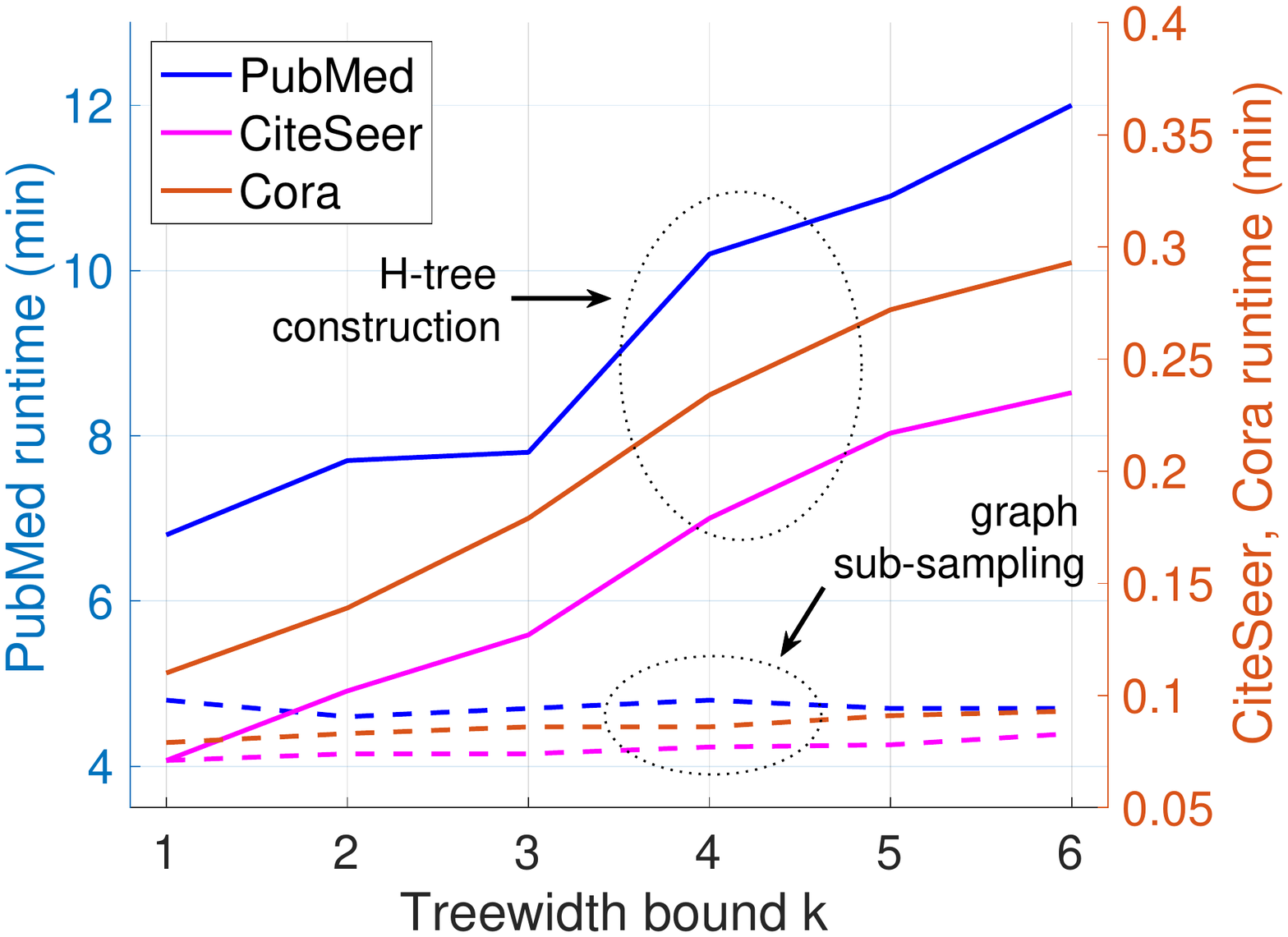}
	\vspace{-0.4cm}
	\caption{Compute time (\Htree and subgraph sampling) as a function of treewidth bound $k$.}
	\label{fig:subsample_and_generate}
\end{wrapfigure}
network datasets.
The reported times are measured when implementing the respective models on an Nvidia Quadro P4000 GPU processor.

Figure~\ref{fig:subsample_and_generate} plots the time required (in minutes) for graph sub-sampling and \Htree construction. We see that while the time required for graph sub-sampling remains nearly the same, the time required for \Htree construction increases in the treewidth bound $k$. This is expected, as for larger $k$, the \Htree construction requires constructing tree-decompositions of many subgraphs of size at most $k$. The absolute numbers reported in Figure~\ref{fig:subsample_and_generate} can be improved as our current implementation uses the popular NetworkX library~\cite{networkX}, which does not produce the time efficient implementation of many of the routines we use. However, we expect the trend observed in Figure~\ref{fig:subsample_and_generate} to hold true.

The increasing compute time with $k$ poses a trade-off between runtime and accuracy, especially for datasets like Cora, where increasing treewidth bound $k$ leads to increase in prediction accuracy.

\end{document}